\documentclass[10pt,journal,compsoc]{IEEEtran}
\ifCLASSOPTIONcompsoc
  \usepackage[nocompress]{cite}
\else
  \usepackage{cite}
\fi
\ifCLASSINFOpdf
\else
\fi

\usepackage{times}
\usepackage{epsfig}
\usepackage{graphicx}
\usepackage{amsmath}
\usepackage{amssymb}
\usepackage{graphics} 
\usepackage{graphicx}
\usepackage{subcaption}
\usepackage{amsthm}

\usepackage{amsmath} 
\usepackage{amssymb}  
\usepackage{color}
\usepackage{mathtools}
\usepackage{bm}
\usepackage{algorithm}
\usepackage{algpseudocode}
\usepackage{multirow}
\usepackage{caption}
\usepackage{booktabs}
\usepackage{dsfont}
\usepackage{array}
\usepackage{enumitem}
\usepackage{url}
\usepackage{cleveref}

\newcommand{\rev}[1]{\textcolor{blue}{#1}}

\setlist[itemize]{noitemsep, nolistsep}

\newcolumntype{P}[1]{>{\raggedright\arraybackslash}p{#1}}

\newcommand{\algName}{FLOWR}
\newcommand{\algNameLong}{few-shot learning for open world recognition}

\newcommand{\algSmallContext}{SC-FLOWR}
\newcommand{\algLargeContext}{LC-FLOWR}

\newcommand{\Ntrain}{N_{\text{train}}}
\newcommand{\Ntest}{N_{\text{test}}}

\newcommand{\Dtrain}{\mathcal{D}_\text{train}}
\newcommand{\Dtest}{\mathcal{D}_\text{test}}
\newcommand{\Dsupp}{\mathcal{D}_\text{supp}}
\newcommand{\Dquer}{\mathcal{D}_\text{query}}
\newcommand{\Dmetatrain}{\mathcal{D}_\text{train}^m}

\newcommand{\citep}[1]{\cite{#1}}
\newcommand{\citet}[1]{\cite{#1}}
\newcommand{\eq}[1]{Eq.~\ref{#1}}

\newcommand{\xdim}{{d_x}}

\newcommand{\phidim}{{d_\phi}}

\newcommand{\tr}{\text{tr}}

\newcommand{\Dquery}{D_\mathrm{query}}

\newcommand{\x}{\bm{x}} 
\newcommand{\y}{\bm{y}} 
\newcommand{\z}{\bm{z}} 

\newcommand{\numclasses}{N}
\newcommand{\numkk}{{N_{kk}}}

\newcommand{\N}{\mathcal{N}}

\definecolor{darkgreen}{rgb}{0,0.5,0}		

\makeatletter
\newcounter{phase}[algorithm]
\newlength{\phaserulewidth}
\newcommand{\setphaserulewidth}{\setlength{\phaserulewidth}}
\newcommand{\phase}[1]{%
  \vspace{-1.25ex}
  \Statex\leavevmode\llap{\rule{\dimexpr\labelwidth+\labelsep}{\phaserulewidth}}\rule{\linewidth}{\phaserulewidth}
  \Statex\strut\refstepcounter{phase}\textit{Phase~\thephase~--~#1}
  \vspace{-1.25ex}\Statex\leavevmode\llap{\rule{\dimexpr\labelwidth+\labelsep}{\phaserulewidth}}\rule{\linewidth}{\phaserulewidth}}
\makeatother

\setphaserulewidth{.2pt}

\usepackage{amsmath,amsfonts,bm}

\def\eqref#1{equation~\ref{#1}}

\def\1{\bm{1}}

\DeclareMathAlphabet{\mathsfit}{\encodingdefault}{\sfdefault}{m}{sl}
\SetMathAlphabet{\mathsfit}{bold}{\encodingdefault}{\sfdefault}{bx}{n}

\newcommand{\R}{\mathbb{R}}

\begin{document}
%
\title{Bayesian Embeddings for Few-Shot\\ Open World Recognition}

\author{John~Willes,
        James~Harrison,
        Ali~Harakeh,
        Chelsea~Finn, 
        Marco~Pavone, 
        and Steven~Waslander%
\IEEEcompsocitemizethanks{\IEEEcompsocthanksitem J.~Willes, A.~Harakeh, and S.~Waslander are with the University of Toronto, Toronto, Ontario, Canada. \protect\\
Email: \texttt{john.willes@mail.utoronto.ca, \hfil\break ali.harakeh@utoronto.ca, stevenw@utias.utoronto.ca} 
\IEEEcompsocthanksitem J.~Harrison, C.~Finn, and M.~Pavone are with Stanford University, Stanford, California, USA.\protect\\
Email: \texttt{\{harrison, cbfinn, pavone\}@stanford.edu}
}
\thanks{Manuscript received April 30, 2021}}

%
%

\markboth{IEEE TRANSACTIONS ON PATTERN ANALYSIS AND MACHINE INTELLIGENCE April~2021}
{Shell \MakeLowercase{\textit{et al.}}: Bare Demo of IEEEtran.cls for Computer Society Journals}

%



\IEEEtitleabstractindextext{%
\begin{abstract}
As autonomous decision-making agents move from narrow operating environments to unstructured worlds, learning systems must move from a closed-world formulation to an open-world and few-shot setting in which agents continuously learn new classes from small amounts of information. This stands in stark contrast to modern machine learning systems that are typically designed with a known set of classes and a large number of examples for each class. In this work we extend embedding-based few-shot learning algorithms to the open-world recognition setting. We combine Bayesian non-parametric class priors with an embedding-based pre-training scheme to yield a highly flexible framework which we refer to as \algNameLong{} (\algName{}).
We benchmark our framework on open-world extensions of the common MiniImageNet and TieredImageNet few-shot learning datasets. Our results show, compared to prior methods, strong classification accuracy performance and up to a 12\% improvement in H-measure (a measure of novel class detection) from our non-parametric open-world few-shot learning scheme.


\end{abstract}

}

\maketitle

\IEEEdisplaynontitleabstractindextext

%
\IEEEpeerreviewmaketitle

\section{Introduction}

The predominant setting for classification systems is \textit{closed-world}: a fixed set of possible labels is specified during training, and this set remains fixed during deployment~\cite{murphy2012machine}. Moreover, modern deep learning classification systems operate in the \textit{big data} regime, in which there are many examples for each class and retraining the classification model is costly in both compute and time. This large-scale, \textit{closed-world} approach stands in stark contrast with human learning in an open world. By constantly integrating novel information, humans continually learn new concepts from small amounts of data. As autonomous decision-making agents move in to increasingly unstructured operating environments---for example, autonomous vehicles being deployed in new cities---learning systems must consider \textit{open-world} and \textit{few-shot} settings in which agents continuously learn new concepts from limited amounts of new information in the wild. 

Recent work has aimed to enable deep learning models to perform few-shot learning, or learning from a small number of examples for each class \cite{finn2017model, vinyals2016matching, snell2017prototypical}. 
These few-shot classification algorithms typically consider the \textit{closed-world} ``$k$-shot, $n$-way'' \cite{vinyals2016matching} setting, in which the number of classes appearing at test time is known, and examples of each class are given in a small training (or context) dataset.
Indeed, work from the last half decade has shown that deep neural network models are capable of learning with limited context data; 
however, these approaches 
fundamentally do not address the problem of continuously detecting and learning from novel classes.  
Moreover, few works have addressed the problem of continuously integrating novel classes over long horizons---a problem setting which spans from few-shot learning to large-scale learning. An effective few-shot, open-world learning pipeline must be able to incorporate arbitrary amounts of training data, detect novel inputs at any point in the evaluation sequence, and instantiate and learn new classes as they appear.


In this work, we propose the small-context and large-context few-shot open-world recognition (FS-OWR) problem settings extending the scope of the existing open-world recognition setting \cite{bendale2015openworld} to include learning with limited labeled data. These settings aim to reflect the deployment of a learning agent into a novel environment, with the difference between the two settings being the amount of directly relevant context data available during training. We argue that these settings are natural for training flexible open-world recognition models, and are thus of rapidly growing importance as classification systems see increasing open world deployment. We present a novel approach to classification in these settings and benchmark algorithms from related domains. Whereas prior work in few-shot learning has addressed parts of this problem statement \cite{ren2019attractor, liu2020openset, ren2020wandering}, we present a differentiable end-to-end framework that is capable of continuously detecting and learning novel classes during operation. Our approach combines ideas from Bayesian non-parametrics~\cite{jordan2015gentle} with a Bayesian formulation of prototypical few-shot learning \cite{snell2017prototypical, harrison2019continuous} to yield a highly flexible and simple non-parametric model capable of reflecting uncertainty in whether a class is novel. In particular, we leverage a Chinese restaurant process (CRP) class prior---a prior on an unbounded number of classes~\cite{gershman2012tutorial}---along with a Bayesian embedding-based meta-learning algorithm~\cite{harrison2019continuous}. To improve the performance of this framework, we present an embedding-based pre-training phase in which a standard fully-connected classification head is replaced with Gaussian class distributions in feature space for pre-training.

\begin{figure*}
\begin{center}
\includegraphics[width=\linewidth]{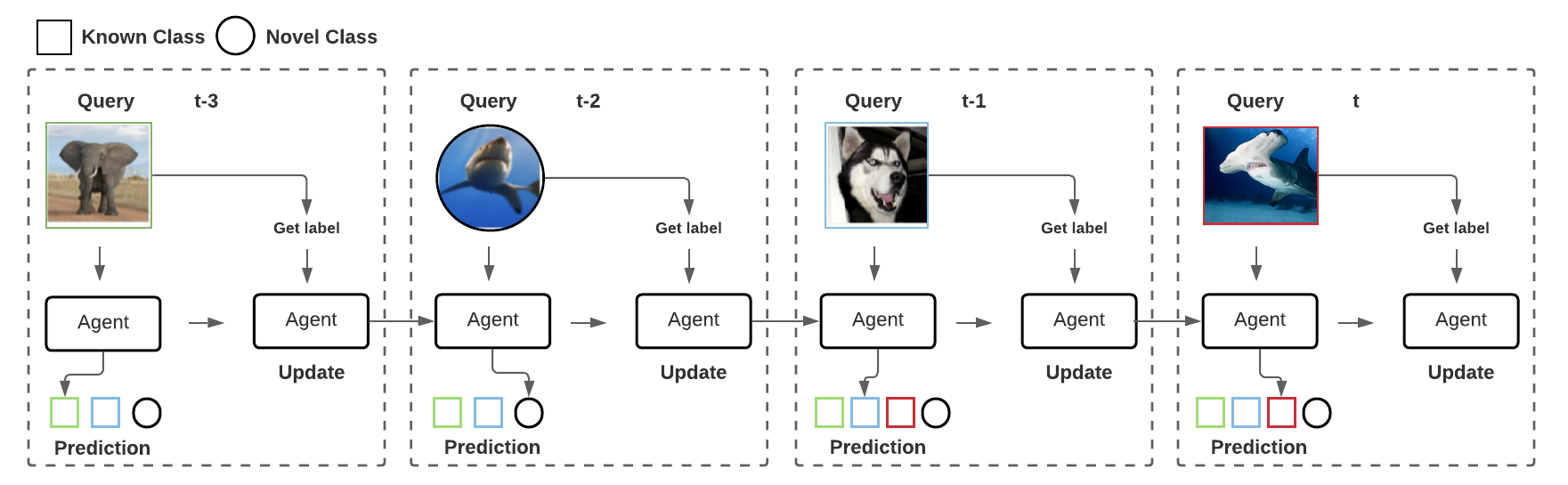}
\end{center}
   \caption{\textbf{Few-shot Open-World Recognition.} In the FS-OWR setting, a learning agent observes a query datapoint at each timestep, $t$. The agent must classify the query as one of a set of known classes or as a novel class (unknown-unknown). Once a prediction is made, the agent receives the true query class label. The new information is used to update its understanding of a known class or instantiate a new known class. In this example, at timestep $t-2$, the agent encounters the novel `shark` class and subsequently increments the set of known classes. The agent must then classify future `shark` queries accordingly, as we see at timestep $t$.  }
\label{fig:fsowr}
\end{figure*}

\subsection{Contributions} There are four core contributions in this paper. 

\begin{itemize}
    \item We propose a formalization of two different few-shot open-world recognition settings in which decision-making agents must classify known classes, detect novel classes and then rapidly adapt and generalize given limited labeled data (Figure \ref{fig:fsowr}). These settings---the small-context and large-context settings respectively---provide a bridge from more narrowly-scoped few-shot learning to a more realistic large-scale continual, open-world learning setting. We adapt and benchmark few-shot learning, open-set recognition and open-world recognition methods to this setting.
    \item We review the use of the area under the receiver operating characteristic (AUROC) as a performance metric for open-world learning algorithms. We demonstrate that AUROC is not model-independent and cannot be use as a reliable metric because it implicitly employs different misclassification cost distributions for different classifiers. We employ the model-independent H-measure~\cite{hand2009hmeasure} as an impartial alternative.
    \item We introduce a Bayesian few-shot learning scheme based on Gaussian embeddings~\cite{snell2017prototypical, liu2020openset}. We combine this approach with a Bayesian non-parametric class prior, and show this system is capable of effectively incorporating novel classes for both few-shot open-world recognition settings. Moreover, we show that our approach results in a 12\% improvement in H-measure for the detection of novel classes when compared to a variety of existing methods when extended to our problem domain. 
    \item Finally, we introduce an embedding-based pre-training phase which better aligns the pre-training model assumptions with those made during meta-learning. This novel pre-training phase has two major impacts: the unified assumptions across training phases enables the large-context problem setting, and the pre-training strategy substantially improves performance across problem settings.
\end{itemize}

\subsection{Organization}

We begin in section \ref{sec:problem} by introducing the few-shot open-world recognition problem setting. In section \ref{sec:related} we discuss related work on open-world learning as well as related problem formulations. Section \ref{sec:GMM} presents the few-shot open-world learning model that forms the basis of our framework, and discusses connections between this model and the current literature on metric-based few-shot learning. In section \ref{sec:approach} we introduce the training details of our approach, as well as test-time considerations. Section \ref{sec:experiments} presents our experimental results on versions of the MiniImageNet \cite{vinyals2016matching,ravi2017optimization} and the TieredImageNet \cite{ren2018meta} datasets. Finally, section \ref{sec:conclusion} presents conclusions as well as future research directions. 

\section{Problem Statement}
\label{sec:problem}

In this work, we aim to develop a classification model that is able to continuously detect novel classes during deployment and incorporate examples of these novel classes to rapidly improve predictive performance.
The evaluation setting is continuous and online: our predictive model is given an image, makes a (probabilistic) label prediction, and then receives the image label.

With this aim in mind, we build on the \textit{open-world recognition} (OWR) setting~\cite{bendale2015openworld}.
Open-world recognition is a multi-task problem consisting of 3 key components: classes that have been seen in prior, offline training must be successfully classified, novel classes must be detected, and recurring novel classes appearing in the online phase must be learned to enable successful classification.
To formalize these goals, we adopt the class taxonomy proposed by~\cite{geng2020recent} as follows\footnote{We narrow the taxonomy to just the elements relevant to few-shot learning. References to side- and attribute-information are omitted because adjacent domains such as zero-shot learning are not within the scope of this work.}: 
\begin{itemize}
    \item \textbf{Known-known classes}, which have labeled data available at training time.
    \item \textbf{Unknown-known classes}, which have no labeled data available at training time, however, labels are available at test time.
    \item \textbf{Unknown-unknown classes}, which have no labeled data available at either training time or test time. Note that in the OWR setting, once a label is provided, unknown-unknown classes will transition to unknown-known classes. 
\end{itemize} 
 A solution must be capable of classifying known-known classes, detecting unknown-unknown classes and incrementally incorporating new information in order to improve classification accuracy for unknown-known classes. A labelling process is also needed for unknown-unknown data. This process may be automated but is typically assumed to be driven by human annotation. Previous work in the open-world recognition setting~\citep{rudd2018evm, bendale2015openworld} makes the assumption that the learning agent has access to a large dataset for known classes. This assumption is not realistic in any situation where the learning agent must adapt rapidly to novel information, and therefore fails the requirements of true open-world learning where the learning agent must be able to integrate new classes given limited labeled data. We argue that the agent should be able to meaningfully learn novel classes from only a small set of labeled examples.


Formally, we propose the \textit{few-shot open-world recognition} (FS-OWR) setting. The FS-OWR setting assumes access to a labeled training dataset $\Dtrain = \{[\bm{x}_1, \ldots, \bm{x}_K], [y_1, \ldots, y_K] \}$, containing $\Ntrain$ classes. Evaluation is performed online using a test dataset $\Dtest = [\x_1, \dots, \x_T]$ of sequence length $T$. $\Dtest$ contains $\Ntest$ classes, where $\Ntrain < \Ntest$. At time $t$, the set of known classes (known-known or unknown-known) is $N_t$, which is initialized as $N_t=\Ntrain$.  The model continuously receives input $\x_t$, and returns a prediction that the input corresponds to a known class or that the input is of an unknown-unknown class. In this work, we will assume this prediction takes the form of a distribution over the $N_t + 1$ classes at time $t$, reflecting the set of known classes plus one class corresponding to a novel input. Following prediction, the model receives the true class label, $y_t$. If the label corresponds to a novel class, the model instantiates a new unknown-known class and increments $N_t$. The task structure has been visualized in Figure \ref{fig:fsowr}. We consider few-shot open-world recognition within two different settings: \textit{small-context} and \textit{large-context}.

\noindent\textbf{Small-Context FS-OWR.} The \textbf{small-context} FS-OWR variant assumes access to a small (or non-existent) $\Dtrain$. This is a similar assumption to that made when sampling tasks in the $k$-shot $n$-way few-shot classification paradigm. Models are evaluated with respect to classification of unknown-known classes and detection of unknown-unknown classes. This variant can be practically motivated by applications in which labels are subjective to a user such as an automated photo tagging service. Users are able to provide labels for images, with which the system learns to automatically classify. Different users may have entirely disjoint sets of labels rendering the collection of a large labelled dataset impossible and detection of inputs outside of the provided set of available labels is critical. Moreover, this setting highlights (or isolates) the few-shot learning element of the problem and thus may be a useful setting within the research community, as the $k$-shot $n$-way setting has been.

\noindent\textbf{Large-Context FS-OWR.} The \textbf{large-context} FS-OWR variant assumes access to a large $\Dtrain$. Performance is measured with respect to classification of known-known classes, unknown-known classes and detection of unknown-unknown classes. We motivate this setting by considering an autonomous vehicle, driving in an urban environment.
This vehicle was trained to recognize classes such as vehicles, cyclists, and pedestrians using a large-scale dataset. However, the set of possible training classes does not encompass the full universe of possible classes observed during deployment. As such, it is desirable that the vehicle is able to recognize entities outside of its training set, so that they may be labeled and the system can adapt.


\begin{figure*}
\begin{center}
\subcaptionbox{\algName{} Prediction\label{pcoc_1}}{
\includegraphics[width=0.32\linewidth]{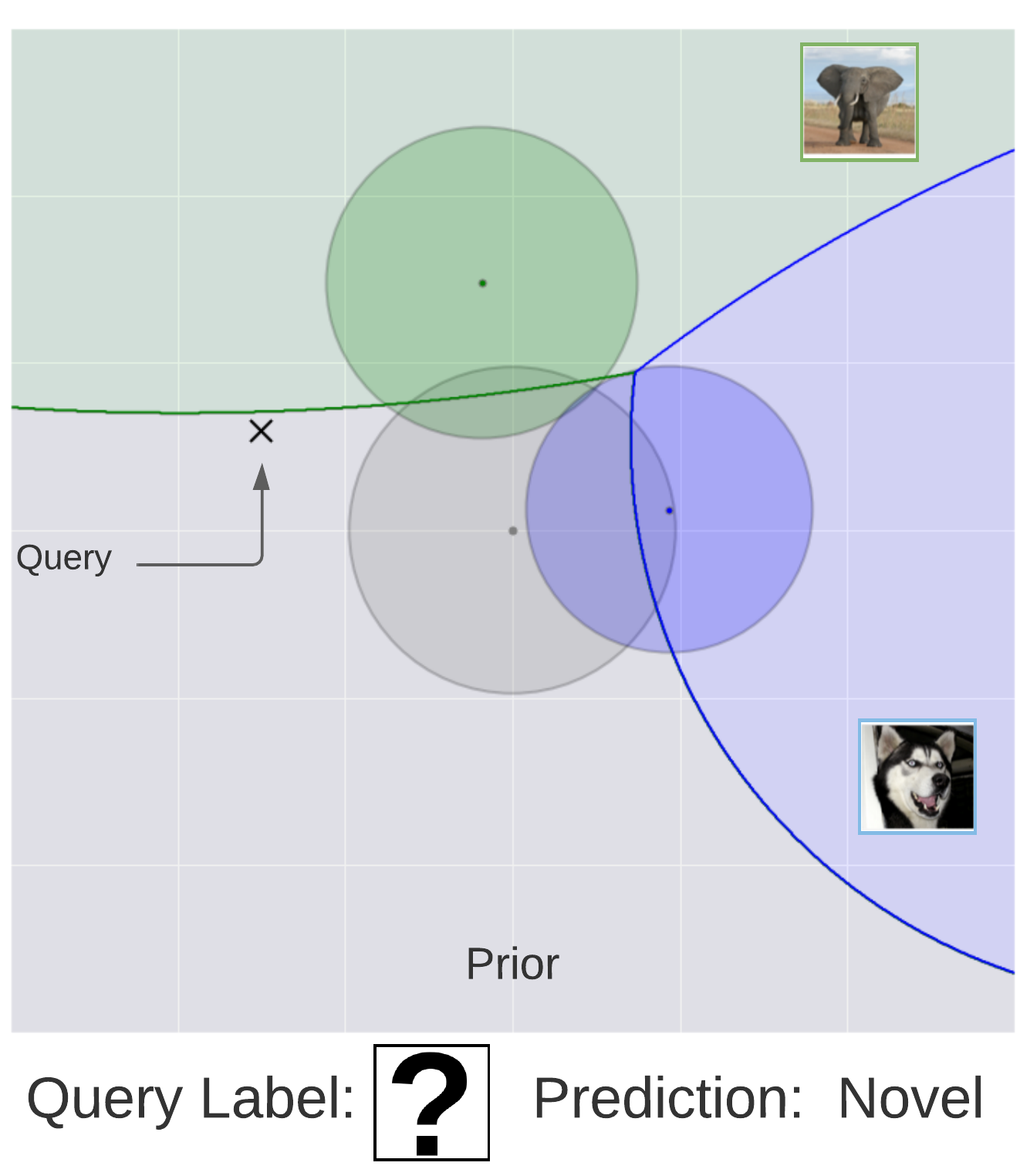}
}%
\subcaptionbox{Novel Class Instantiation\label{pcoc_2}}{
\includegraphics[width=0.32\linewidth]{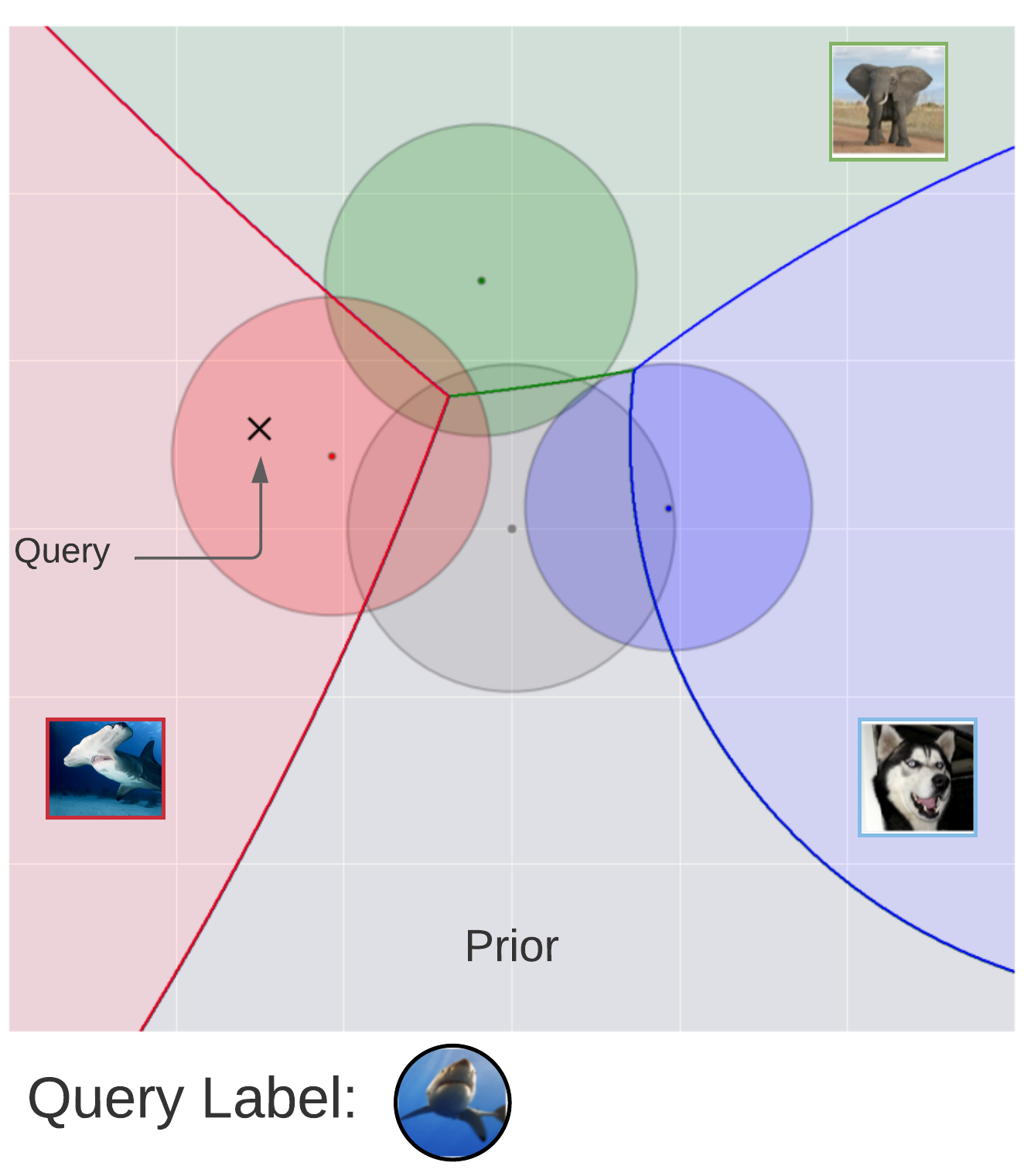}
}
\subcaptionbox{Known-Class Update\label{pcoc_3}}{
\includegraphics[width=0.32\linewidth]{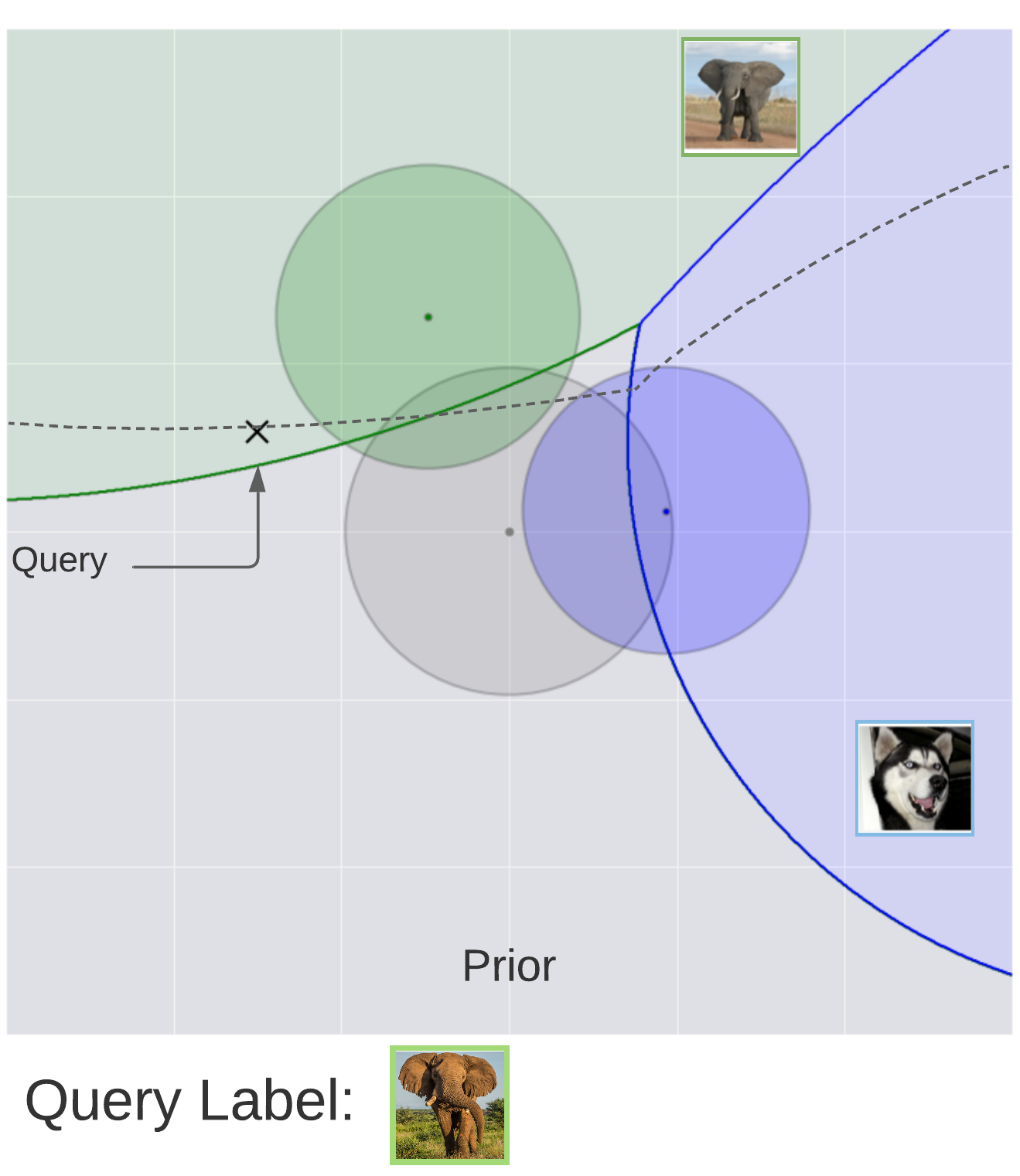}
}
\end{center}
   \caption{A 2D visualization of \algName{} decision boundaries and adaptation. The shared prior and its decision boundary corresponding to a novel class are depicted in gray. The known class posterior predictive distributions and their associated decision boundaries are colored blue, green and red. Each circle depicts the $2\sigma$ confidence interval of an isotropic Gaussian. The black ``X" denotes a query feature vector. In this case (fig. \ref{pcoc_1}),  \algName{} classifies the query as a novel class.  \algName{} will then receive the true class label. If the label corresponds to a novel class (fig. \ref{pcoc_2}), \algName{} will instantiate a new class distribution from the shared prior and condition on the query feature (shown in red). In the case that label corresponds a known-class (fig. \ref{pcoc_3}), \algName{} will update the known-class posterior predictive to extend the decision boundary to include the query. The previous decision boundaries are depicted with a dashed line.}
\label{fig:lpcoc}
\end{figure*}

\section{Related Work}
\label{sec:related}

Open-world classifiers \citep{rudd2018evm, bendale2015openworld} must be capable of rejecting query data that is not supported by known class labeled data. These classifiers typically rely on a minimum-threshold of the known class probabilities in order to manage this open space risk \cite{geng2020recent}. The Extreme Value Machine (EVM) \cite{rudd2018evm} is a non-parametric classifier which leverages statistical extreme value theory \cite{fisher1928limiting} to fit distributions to the margin distances of labeled data. Class probabilities are thresholded to define a boundary between known classes and unsupported open-space. While thresholding provides a convenient means of bounding the open-space it is non-differentiable and is challenging to reconcile with modern gradient-based machine learning techniques, especially in an open-world setting. The meta-learned prior proposed in this work provides an alternative which can be learned end-to-end. 

Few-shot learning has seen increasing attention in recent years due to the difficulty of learning from a small number of examples with deep neural network models \cite{ravi2017optimization, snell2017prototypical}. Meta-learning---or ``learning-to-learn''---has been shown to be an effective approach to the few-shot learning problem \cite{finn2017model, vinyals2016matching, snell2017prototypical}. By training on many different few-shot learning problems, an inner adaptation process is learned, capable of rapid learning given few examples. This primarily takes the form of learning an initialization or prior for a learning algorithm, an update rule to incorporate data, or both.  

Although there are many taxonomies of meta-learning algorithms \citep{hospedales2020meta,vanschoren2018meta}, three common overarching categories have emerged: recurrence-based methods, optimization-based methods, and metric based-methods. Recurrence-based methods largely focus on a black-box update procedure; the inner learning algorithm typically takes the form of a recurrent neural network \citep{hochreiter2001learning}. The black-box descriptor highlights the fact that the inner learning algorithm does not leverage any inductive biases from e.g. optimization. Optimization-based meta-learning aims to explicitly use an optimization procedure in the inner learning algorithm, and typically outperforms black-box methods as a result \cite{finn2017model}. These approaches rely on back-propagating through an optimization problem, yielding a bi-level optimization problem. Common approaches include back-propagation through the sequence of gradient updates for a set of parameters, as in MAML \citep{finn2017model}, or back-propagation through the fixed point of a convex optimization algorithm \citep{harrison2018meta, bertinetto2018meta, rajeswaran2019meta, lee2019meta}. Finally, metric-based meta-learners rely on the inductive bias of metric learning, in which nearby samples in an embedding space are likely to be of the same class. Generally, these methods aim to learn an embedding space and/or a metric in this space, such that the embeddings of inputs of the same class are close to each other, and different classes are separable based on the metric \citep{snell2017prototypical}. 

Of particular interest are recent works investigating meta-learning for incremental learning or open-set recognition. As described in Section \ref{sec:problem}, both of these settings are strongly related to the open-world recognition setting. Incremental learning corresponds to an expansion of the possible inputs \citep{castro2018end, rebuffi2017icarl}. In particular, in the classification setting, incremental learning methods typically assume the addition of a new class, and must learn to correctly classify inputs corresponding to this class without needing to detect unknown-unknown class data. Open-set recognition \cite{bendale2016towards, geng2020recent} focuses on the complementary scenario, in which a classifier must classify known classes and detect unknown-unknown class data with needing to incorporate detected novel classes. \cite{santoro2016meta} meta-train memory-augmented neural networks to incorporate incremental classes across few-shot tasks but do not detect novel classes. \cite{liu2020openset} investigate open-set meta-learning; their approach is based on thresholding on prototypical embeddings. Because they lack the prior we introduce in this work, there is no clear method to instantiate new classes. A similar approach was developed by \cite{allen2019infinite}, who base their approach on a discrete clustering algorithm that can be seen as the limiting case of the CRP prior.\cite{ren2019attractor} consider a few-shot approach to incremental learning in which a base classifier is paired with a standard few-shot learner. To classify incremental classes, the authors introduce an attention-based mechanism, trained via meta-learning. 

Other related settings include ``online contextualized few-shot learning'' \cite{ren2020wandering} and ``in the wild''\cite{wallingford2020wild}.  ``Online contextualized few-shot learning'' proposes a setting in which classes grow incrementally as they are observed and novel inputs are detected. Unlike this work, there is no distinction between model training and testing and there is no means to exploit large-scale datasets for known-known classification. ``In the wild'' present a unified setting for incremental, and few-shot learning. They benchmark performance for several optimization and embedding-based few-shot learning schemes, with and without pre-training. This work showed the importance of non-meta-learning pre-training and also demonstrated that simple embedding-based methods are highly competitive with other investigated approaches. 

\section{The Non-Parametric Gaussian-Dirichlet Mixture Model}
\label{sec:GMM}

In this section we describe our non-parametric Gaussian-Dirichlet Mixture Model that lies at the core of the \algName{} framework (visualized in Figure \ref{fig:lpcoc}). We first describe at a high level our formulation of few-shot learning as Bayesian inference in a mixture model. Following this, we describe each necessary sub-component in detail. Finally, we discuss in more detail the literature on metric-based few-shot learning, and how our approach in this section relates to that prior work. 

\subsection{Open-World Few-Shot Learning}

Our approach builds on a line of work in few-shot learning/meta-learning as Bayesian inference \cite{grant2018recasting, harrison2018meta, gordon2018meta, harrison2019continuous}. In particular, let $X = [\bm{x}_1, \ldots, \bm{x}_K], Y = [y_1, \ldots, y_K]$ denote context data, used to improve predictions. Practically, context data is the complete set of labeled data available (either known-known or unknown-known class data). We assume $y_k \in \mathbb{N}^+$ for all $k$. We wish to optimize the posterior predictive likelihood\footnote{We assume independence between the parameters of the generative model of $y$ and those of the conditional generative model of $y\mid \bm{x}$, and thus we have $p(y_*\mid Y,X) = p(y_*\mid Y)$.}
\begin{equation}
    p(y_* \mid X, Y, \bm{x}_*) = \frac{p(\bm{x}_*\mid X, Y, y_*) p(y_*  \mid Y)}{ p(\bm{x}_*\mid X, Y)}
\end{equation}
in expectation over \textit{context data}\footnote{This process is equivalent to optimizing the marginal likelihood in empirical Bayesian inference \cite{grant2018recasting,robbins1956empirical}.} $X,Y$ and test datapoint $\bm{x}_*, y_*$. Critically, our approach must be capable of handling a growing number of classes to handle the open-world setting. We turn to a non-parametric embedding-based few-shot learning framework based on the Dirichlet process Gaussian mixture model \cite{neal2000markov, fox2009bayesian}.

Following prior work in metric-based few-shot learning \cite{snell2017prototypical}, we map our inputs through an \textit{encoder network} $\bm{\phi}:\R^\xdim \to \R^\phidim$. We write $\bm{z}_k = \bm{\phi}(\bm{x}_k)$. We then perform Gaussian mixture model-based generative modeling in the output space of this encoder. By leveraging a Dirichlet process \cite{gershman2012tutorial, jordan2015gentle} class prior ${p(y_*  \mid Y)}$ combined with a generic prior over embedding means, this model is capable of incorporating new classes. More precisely, we assume a generative model in feature space of the form
\begin{equation}
\begin{aligned}
    \bm{p} &\sim \text{DP}(\bm{c}_0)\\
    y &\sim \text{Cat}(\bm{p})\\  
    \bm{z} \mid y = n &\sim \N(\bar{\bm{z}}[n], \Sigma_{\epsilon}) \\
    \bar{\bm{z}}[n] &\sim \N(\bm{\mu}_0, \Sigma_0)
\end{aligned}
\end{equation}
where $\text{DP}(\cdot)$ denotes a Dirichlet process with hyperparameters $\bm{c}_0$, which is a distribution over distributions whose realizations are themselves probability measures. We use square brackets to refer to a particular class index. The term $\bar{\bm{z}}$ refers to the set of all $\bar{\bm{z}}[n]$, where ${n \leq \numclasses = \text{max}(y_k)}$. 



\subsection{Priors and Posterior Inference}

Generally, inference in the unsupervised Dirichlet process Gaussian mixture model is analytically intractable, and must turn to MCMC sampling methods or variational inference methods \cite{neal2000markov}. However, we consider a fully supervised setting, visualized in Figure \ref{fig:fsowr}, in which after a prediction for the label of an input $\bm{x}_*$ is made, we receive the label $y_*$. In our inference model, we will compute posterior predictives for $\bm{z}_*$---these quantities are useful for computing the posterior predictive over the label, but we emphasize that we do not build a generative model of the image inputs, but one for learned features relevant for classification.

\subsubsection{Gaussian Posterior Inference}

We first look at computing the class conditional posterior ${p(\bm{z}_*\mid X, Y, y_*)}$. 
We write 
$\bm{z}_* = \bm{\phi}(\bm{x}_*)$. 
We will write the mean for class $n$ as $\bar{\bm{z}}[n]$, and we place a Gaussian prior over $\bar{\bm{z}}[n]$ with mean $\bm{\mu}_0$ and covariance $\Sigma_0$. This prior is shared for all classes, which is central to the ability of our model to instantiate new classes. Given our Gaussian prior on $\bar{\bm{z}}[n]$ and the assumed Gaussian likelihood, the posterior over $\bar{\bm{z}}[n]$ is analytically tractable due to Gaussian-Gaussian conjugacy \cite{murphy2012machine}. The posterior after conditioning on $X,Y$ is of the form 
\begin{equation}
    p(\bar{\bm{z}}[n] \mid X,Y) = \N(\bm{\mu}_K[n], \Sigma_K[n])
\end{equation}
where the subscripts denotes the posterior after conditioning on $K$ datapoints. We will describe the computation of the posterior statistics for this model via the natural parameters:
\begin{equation}
    \label{eq:pcoc-factorization}
    \begin{aligned}
        \bm{q}_K[n] &= \Sigma^{-1}_K[n] \bm{\mu}_K[n]\\
        \Lambda_K[n] &= \Sigma^{-1}_K[n].
    \end{aligned}
\end{equation}
The update equations for posterior parameters given data $(\bm{x}_{k+1}, y_{k+1} = n)$ are written recursively as
\begin{equation}\label{eq:pcoc-recursive}
\begin{aligned}
    \bm{q}_{k+1}[n] &= \bm{q}_{k}[n] + \Sigma_{\epsilon}^{-1} \bm{\phi}(\bm{x}_{k})\\ 
    \Lambda_{k+1}[n] &= \Lambda_{k}[n] + \Sigma^{-1}_{\epsilon}
\end{aligned}
\end{equation}
with $\bm{q}_{k+1}[m] = \bm{q}_{k}[m]$ and $\Lambda_{k+1}[m] = \Lambda_{k}[m]$ for $m \neq n$.
These quantities are initialized as 
\begin{equation}
    \begin{aligned}
    \bm{q}_0[n] &= \Sigma_0^{-1} \bm{\mu}_0 \quad &&\forall n = 1, \ldots, \numclasses\\
    \Lambda_0[n] &= \Sigma_0^{-1}  \quad &&\forall n = 1, \ldots, \numclasses.
    \end{aligned}
\end{equation}

Given these posterior statistics, we may compute the desired predictive distribution via marginalizing
\begin{equation}
    p(\bm{z}_* \mid X, Y, y_*) = \int p(\bm{z}_* \mid \bar{\bm{z}}[n], y_*) \, p(\bar{\bm{z}}[n] \mid X, Y) d\bar{\bm{z}}[n]
\end{equation}
which is analytically tractable, yielding posterior predictive 
\begin{equation}
    p(\bm{z}_* \mid X, Y, \bm{y}_* = n) = \N(\bm{\mu}_K[n], \Sigma_K[n] + \Sigma_\epsilon).
\end{equation}



\subsubsection{Computing the Class Prior}

In this subsection we discuss the computation of $p(y_*  \mid Y)$. We place a Dirichlet process prior on $y$ that allows for an unbounded number of classes. In particular, we leverage the Chinese restaurant process (CRP) which has been effectively used in other works \citep{nagabandi2019deep, allen2019infinite, jerfel2019reconciling}. For a more detailed description than the brief outline presented here, we refer the reader to \cite{gershman2012tutorial,orbanz2010bayesian, jordan2015gentle}.

The standard CRP has two concentration parameters, $a$ and $b$. Assume we have observed $k$ datapoints belonging to classes $n = 1, \ldots, N$. Let $k_n$ denote the number of datapoints in class $n$. Then, the posterior over datapoint $(\bm{x}_{k+1}, y_{k+1})$ belonging to class $N + 1$ (without having conditioned on $\bm{x}_{k+1}$) is 
\begin{equation}
\label{eq:crp_novel}
    p(\y_{k+1} = N+1) = \frac{a + b N}{k + b}
\end{equation}
and the probability of belonging to class $n$ is 
\begin{equation}
\label{eq:crp_base}
    p(\y_{k+1} = n) = \frac{k_n - a}{k + b}.
\end{equation}
where $a \in [0, 1]$ and $b > - a$ to ensure the distribution is a valid probability measure. In practice, we find that learning both CRP parameters can lead to instability during training. Therefore, we treat $a$ as a hyper-parameter, while $b$ is implemented as the sum of $-a$ and a strictly positive term enforced using a softplus activation function.

As can be seen from \eq{eq:crp_novel} and \eq{eq:crp_base}, inference in this model is similar to Dirichlet distribution, with the sufficient statistics corresponding to simple counts. We initialize a vector $\bm{c}_0 = [1]$, and update it as
\begin{equation}
    \begin{aligned}
    \bm{c}_{k+1}[y_k] &= \bm{c}_{k}[y_k] + 1\\
    \bm{c}_{k+1}[m] &= \bm{c}_{k}[m], \quad \forall m \neq y_k\\
    \end{aligned}
\end{equation}
if $y_{k+1} \leq \text{len}(\bm{c})$ and 
\begin{equation}
    \bm{c}_{k+1} = [\bm{c}_{k}, 1]
\end{equation}
otherwise, where $[\cdot,\cdot]$ denotes concatenation. Thus, the CRP prior provides a convenient mechanism to both predict novel classes with tunable probability (depending on $a,b$) and instantiate new class embeddings when a novel class occurs. 

\subsubsection{Computing the Posterior Predictive}
\label{sec:computing-posterior-predictive}

As noted at the beginning of this section, our exact Bayesian inference formulation enables us to make a prediction for test point $\bm{z}_* = \bm{\phi}(\bm{x}_*)$ using Bayes rule, 
\begin{equation}
    \label{eq:prediction}
    p(y_* \mid X, Y, \bm{z}_*) = \frac{p(\bm{z}_*\mid X, Y, y_*) p(y_*  \mid Y)}{ p(\bm{z}_*\mid X, Y)}.
\end{equation}
In the previous two subsections we have presented a methodology for computation of these quantities. We will again assume after observing $X,Y$ that we have seen $\numclasses$ classes. Due to $y_*$ necessarily being discrete, 
\begin{equation}
    p(\bm{z}_*\mid X, Y) = \sum_{y=1}^{N+1} p(\bm{z}_*\mid X, Y, y) p(y \mid Y)
\end{equation}
is analytically tractable. Thus, every step of the procedure required to compute the posterior predictive may be done analytically due to conjugacy. This is critical to enable both efficient and differentiable inference. This differentiability is leveraged in the next section to enable meta-training of this model. 

We note that Bayes' rule requires marginalization over the evidence. Within our model, this is done by applying a softmax to the unnormalized class logits~\cite{jordan1995logistic}, which is similar to the standard design of discriminative neural network classifiers. Therefore, the unnormalized class logits are computed as the log posterior predictives and the prior evaluated at $\z_*$. Application of the softmax performs the marginalization as follows:

\begin{equation}
    p(y_* \mid X, Y, \bm{z}_*) = \operatornamewithlimits{softmax}_{n=1,\ldots,N+1}(\log p(\bm{z}_*, y[n] \mid X, Y))
\end{equation}
where for all known classes we compute

\begin{equation}
   p(\bm{z}_*, y[n] \mid X, Y) = \N(\bm{z}_* \mathbf{;} \ \bm{\mu}_K[n], \Sigma_K[n] + \Sigma_\epsilon)\left(\frac{k_n - a}{k + b}\right)
\end{equation}
and for the novel class we compute
\begin{equation}
    p(\bm{z}_*, y[n] \mid X, Y) = \N(\bm{z}_* \mathbf{;} \ \bm{\mu}_0, \Sigma_0 + \Sigma_\epsilon)\left(\frac{a + b N}{k + b}\right).
\end{equation}

We discuss training the encoder network in detail in the next section. Learning both the shared prior and the noise covariance, $\Sigma_\epsilon$, is redundant when both are parameterized with the same sparsity pattern. 
We therefore fix the class noise covariance  to be the same for all classes, which we write as $\Sigma_\epsilon$ for all $n \in {1,\ldots,N}$ and treat it as a hyperparameter in training.  In practice we fix $\Sigma_\epsilon$ to be isotropic; this assumption could be relaxed to a diagonal covariance with little increase in computational complexity. However, moving to a dense covariance matrix results in cubic complexity for required matrix inversions compared to linear complexity for a diagonal covariance, and so we leave further investigation of dense covariance parameterizations to future work (we refer the reader to \cite{harrison2021uncertainty} for further discussion). We found no substantial performance differences between diagonal and isotropic covariance matrices in our experiments. As we discuss in section \ref{sec:ft}, we perform an additional last layer fine tuning step that adapts the representation toward a more spherical covariance.


\subsection{Discussion}

Our model builds on a line of metric-based meta-learners. Prototypical networks \cite{snell2017prototypical}, the most popular metric-based meta-learner, is also potentially the simplest. In the prototypical networks approach, class prototypes $\bm{r}_n$ are defined as the mean features for a given class,
\begin{equation}
    \bm{r}_n = \frac{\sum_{\bm{x}\in X, y\in Y} \bm{\phi}(\bm{x}) \mathbb{I}(y = n)}{\sum_{y\in Y} \mathbb{I}(y = n)} 
\end{equation}
where $\mathbb{I}(\cdot)$ denotes the indicator function that evaluates to 1 if the input is true and otherwise evaluates to zero. Then, the predicted distribution over labels for a novel test class is
\begin{equation}
    p(y_* = n \mid \bm{x}_*) = \frac{\exp(-\|\phi(\bm{x}_*) - \bm{r}_n \|^2_2)}{\sum_{n'=1}^\numclasses \exp(-\|\phi(\bm{x}_*) - \bm{r}_{n'} \|^2_2)}.
\end{equation}
This predictive likelihood may be used to train the neural network features. The inner learning procedure corresponds to linear discriminant analysis \citep{murphy2012machine} with an isotropic covariance. This interpretation, in which in the inner learning algorithm is a Gaussian discriminant analysis learning algorithm and the outer loop learns neural network features, has spawned numerous extensions. 

Ren et al. \cite{ren2018meta} extend the model with both semi-supervised learning and learned covariances, and \cite{allen2019infinite} extend the approach with a clustering framework derived from the limit of the Dirichlet process. Most related to our approach is PCOC \cite{harrison2019continuous}, who also leverage the conjugacy of the Gaussian-Gaussian mixture model to perform exact Bayesian inference. There are two major differences between this work and ours. Firstly, whereas PCOC uses a Dirichlet distribution class prior, we extend this prior to a Dirichlet process, enabling open-world learning. 
Secondly, PCOC learns a Gaussian prior over each class mean for a fixed number of classes. This is a common approach in few-shot learning; MAML \citep{finn2017model} makes a similar assumption when defining the dimension of the final classification network layer. In both cases, the model adapts at test-time to unknown-known class support data and the class indices have a semantic meaning. However, these models are limited to closed-world classification tasks of a fixed size. In this work we instead learn a single shared Gaussian prior over the mean of all class embeddings, which has mean $\bar{\bm{z}}_0$ and covariance $\Lambda_0$. 
The advantages of the shared prior are two-fold. First, the parameters of the prior distribution can be used to instantiate novel classes by initializing the recursive update equations for the statistics of the mean and variance of class embedding means, $\bm{q}[n]$ and $\bm{\Lambda}[n]$. The details of this recursion are provided in Algorithm~\ref{algo:update}. Second, the prior also provides a distribution for samples from unknown-unknown classes and provides a bound on the open-space for known classes.

\begin{algorithm}[t] 
\begin{algorithmic}[1]
    \Require{$\alpha, \beta, \Dtrain$}
    \State $\numkk = \max(y)$ for $y \in \Dtrain$ \Comment{Number of training classes}
    \State Randomly initialize: $\bm{\phi}$
    \State Randomly initialize: $\bm{\mu}_e \leftarrow [\bm{\mu}_1, \dots, \bm{\mu}_\numkk]$
    \State Randomly initialize: $\bm{\Sigma}_e \leftarrow [\bm{\Sigma}_1, \dots, \bm{\Sigma}_\numkk]$
    \While{not done}
    \State Sample mini-batch: $\mathcal{B}_i \sim \Dtrain$
    \State Evaluate: $\mathcal{L} \leftarrow \mathcal{L}_{\text{NLL}}(\bm{\phi}, \bm{\mu}_e, \bm{\Sigma}_e, \mathcal{B}_i) + \beta \sum_{n=1}^{\numkk} \tr(\bm{\Sigma}_n^{-1})$
    \State Update: $(\bm{\phi}, \bm{\mu}_e, \bm{\Sigma}_e) \leftarrow (\bm{\phi}, \bm{\mu}_e, \bm{\Sigma}_e) - \alpha \nabla \mathcal{L}$
    \EndWhile
    \State \Return $(\bm{\phi}, \bm{\mu}_e, \bm{\Sigma}_e)$
\end{algorithmic}
 \caption{Supervised-Embedding Pre-Training}
 \label{algo:pretraining}
\end{algorithm}

\section{\algNameLong{}}
\label{sec:approach}

Our training pipeline relies on two primary phases. We first pre-train the encoder, which has been shown to substantially improve performance of meta-learners in the few-shot setting \citep{chen2020new, triantafillou2019meta}. The second phase consists of a meta-training phase, either in the small-context setting or large-context setting. A test-time fine-tuning of our encoder is also performed in the small-context setting. We will first describe the pre-training approach before introducing the \algNameLong{} (\algName{}) framework. We will then describe the meta-training procedure in both the small- and large-context setting, as well as the fine tuning method.

\begin{algorithm}[t] 
\begin{algorithmic}[1]
    \Require{$\bm{z}, y, \bm{c}, \bm{q}, \bm{\Lambda}, \numkk$}
    \State $\numclasses = \text{len}(\bm{c})$ \Comment{Number of observed classes}
    \If{$y = \numclasses + 1$} \Comment{Novel Class}
        \State Append: $\bm{c} \leftarrow [\bm{c}, 1]$
        \State Append: $\bm{q} \leftarrow [\bm{q}, \bm{q}_0]$ 
        \State Append: $\bm{\Lambda} \leftarrow [\bm{\Lambda},  \Lambda_0]$
    \EndIf
    
    \State Update: $\bm{c}[y] \leftarrow \bm{c}[y] + 1$
    
    \If{$y > \numkk$} \Comment{Update Unknown-Known Class}
        \State Update $\bm{q}[y] \leftarrow \bm{q}[y] + \Sigma_\epsilon^{-1} \bm{z}$
        \State Update $\bm{\Lambda}[y] \leftarrow \bm{\Lambda}[y] + \Sigma_\epsilon^{-1}$
    \EndIf
    
    \State \Return $\bm{c}, \bm{q}, \bm{\Lambda}$
\end{algorithmic}
 \caption{Update}
 \label{algo:update}
\end{algorithm}

\subsection{Pre-Training}

We adapt the methodology of \citep{chen2020new} to pre-train an encoder before meta-training. In \cite{chen2020new} pre-training follows standard large-scale classification, wherein a neural network is transformed by a fully connected linear last layer to a set of logits (unnormalized probabilities), and then passed through a softmax function to normalize. These models result in polytopic regions in feature space, with potentially high predicted class probability far from training data. However, the geometry of the classification regions for this pre-training are likely extremely different to that of the Gaussian embedding-based model presented in the previous section, which naturally decay to high uncertainty far from training data \cite{lee2018simple}.

To address this mismatch in classifier confidence, our pretraining learns Gaussian class embeddings in feature space that allow for efficient initialization of our meta-learning model in both the SC and LC settings. Our supervised embedding (Sup-E) approach is described in Algorithm \ref{algo:pretraining}. We write $\bm{\phi}$ to flexibly describe both the encoder function as well as the weights of this neural network, depending on context. We assume a uniform distribution\footnote{The assumption of uniform class probabilities may easily be relaxed.} over classes and directly learn a set of Gaussian embeddings using a large-scale training set, $\Dtrain$, containing $\numkk=\max(y)$ for $y \in \Dtrain$ classes. The embeddings consist of a set of class means, $\bm{\mu}_e = [\bm{\mu}_1,\dots, \bm{\mu}_\numkk]$, and covariances, $\bm{\Sigma}_e = [\Sigma_1, \dots, \Sigma_\numkk]$, which are learnable network parameters trained using a negative predictive log-likelihood (NLL) loss function. The covariance is constrained to be isotropic and inference is performed via Gaussian discriminant analysis. Prediction is similar to the procedure described in section \ref{sec:computing-posterior-predictive}. Class probabilities are computed by evaluating each of the learned Gaussian embeddings at $\z_*$ followed by a softmax:

\begin{equation}
    p(y_* \mid X, Y, \bm{z}_*) = \operatornamewithlimits{softmax}_{n=1,\ldots,\numkk}(\log \N(\z_* \mathbf{;} \ \bm{\mu}_n, \Sigma_n)).
\end{equation}
As our approach relies on Gaussian embedding methods, there is better agreement between the pre-training and the meta-training phase under Gaussian embedding-based pre-training. 

The class embeddings are regularized via a Gaussian prior on the mean and an inverse Wishart prior on the covariance. Thus, the learned point estimate of the mean and covariance of the embeddings may be seen as the MAP estimates through the lens of Bayesian inference. These priors correspond to a simple $L_2$ regularization on the mean of the classes (corresponding to a prior centered on the origin), and a combination of a trace regularization term and a log determinant regularization term on the inverse covariance. In practice we only impose the trace regularization term---corresponding to simply penalizing the inverse of the variance scale in the isotropic case---as the trace term is the primary factor preventing highly concentrated embeddings. Our complete pre-training loss function is therefore:

\begin{equation}
\label{eq:pretrain-loss}
    \mathcal{L} = \mathcal{L}_\text{NLL} + \beta \sum_{n=1}^{\numkk} \tr(\Sigma_n^{-1})
\end{equation}
where the strength of this regularization, $\beta$, was determined by performance on the validation set. We note that we regularize the embeddings toward the same mean prior used in the meta-training phase, thus strengthening the connections between the pre-training and meta-training phases. However, the strength of the regularization of the covariance is arbitrary, as we do not try to infer the noise covariance in the meta-learning pipeline, and instead treat it as a point estimate that is known to the inner learning algorithm.

\begin{algorithm}[t] 
\begin{algorithmic}[1]
    \Require{$\bm{\phi}, \bm{q}_0, \Lambda_0, \Sigma_\epsilon, a, b, \Dsupp$}
    \phase{Initialize}
    \State Initialize class counts: $\bm{c} \leftarrow [\,]$ 
    \State Initialize: $\bm{q} \leftarrow [\bm{q}_0\,]$ 
    \State Initialize: $\bm{\Lambda} \leftarrow [\Lambda_0\,]$ 
    \State $\numkk = 0$ \Comment{No known-known classes}
    \For{$(\bm{x}, y)$ in $\Dsupp$} 
      \State $\text{Update}(\bm{\phi}(\bm{x}), y, \bm{c}, \bm{q}, \bm{\Lambda}, \numkk)$ \Comment{Algorithm \ref{algo:update}}
    \EndFor
    \phase{Predict \& Update}
    \While{not done}
        \State Given query datapoint: $\bm{x}_{*}$
        \State $\hat{\bm{y}}$ = $\text{Predict}(\bm{\phi}(\bm{x}_{*}), \bm{q}, \bm{\Lambda}, \bm{c}, a, b)$ \Comment{Eq. \ref{eq:prediction}} 
        \State Given query class label: $y_{*}$
        \State $\text{Update}(\bm{\phi}(\bm{x}_*), y_*, \bm{c}, \bm{q}, \bm{\Lambda}, \numkk)$ \Comment{Algorithm \ref{algo:update}}
    \EndWhile
\end{algorithmic}
 \caption{Small-Context \algName{} Prediction \& Update}
 \label{algo:sc-owpcoc}
\end{algorithm}

\subsection{Small and Large-Context \algName}

We now present an overview of the online (test-time) behavior of our \algName{} framework, in both the small-context (SC) and large-context (LC) setting. 

\subsubsection{Small-Context \algName{} (\algSmallContext{})}
As described in Section \ref{sec:problem}, the small-context setting considers unknown-known and unknown-unknown class data. No knowledge of test-time classes is available during training, necessitating the instantiation of all class distributions via the shared prior by conditioning on unknown-known class data. Our meta-training and pre-training phase thus rely on a dataset which we assume consists of different classes. We initialize \algSmallContext{} in the small-context setting given shared prior statistics ($\bm{q}_0, \Lambda_0, \Sigma_\epsilon)$, CRP statistics $(a, b)$, and a small labeled support set (training set), $\Dsupp$, containing unknown-known class data. We initialize the set of class statistics from the shared prior as $\bm{q}$ and $\bm{\Lambda}$ and compute the posterior statistics using the recursive update equations (Algorithm \ref{algo:update}) for all $(\bm{x}, y) \in \Dsupp$. The CRP $\bm{c}$ class counts are maintained simultaneously. As query datapoints, $\bm{x}_*$, are observed, predictions are performed via Equation \ref{eq:prediction}. Once a prediction is made, the model receives a label, $y_*$, indicating that the query belongs to a known class or unknown-unknown class ($y_* = \numclasses + 1$). We will also refer to these unknown-unknown classes as novel classes. In the case of an unknown-unknown class label the model will instantiate a new known class using the shared prior and increment the number of known classes, $\numclasses$. In either case, the model will then condition on the new labeled datapoint. The complete procedure is provided in Algorithm \ref{algo:sc-owpcoc}.

\begin{algorithm}[t] 
\begin{algorithmic}[1]
    \Require{$\bm{\phi}, \bm{q}, \bm{\Lambda}, \Sigma_\epsilon, a, b, \numkk$}
    \phase{Initialize}
    \State Initialize CRP class counts: $\bm{c} \leftarrow [0,...,0]$ \Comment{$\text{Len}(\bm{c}) = \numkk$}
    
    \phase{Predict \& Update}
    \While{not done}
        \State Given query datapoint: $\bm{x}_{*}$
        \State $\hat{\bm{y}}$ = $\text{Predict}(\bm{\phi}(\bm{x}_{*}), \bm{q}, \bm{\Lambda}, \bm{c}, a, b)$ \Comment{Eq. \ref{eq:prediction}} 
        \State Given query class label: $y_{*}$
        \State $\text{Update}(\bm{\phi}(\bm{x}_*), y_*, \bm{c}, \bm{q}, \bm{\Lambda}, \numkk)$ \Comment{Algorithm \ref{algo:update}}
    \EndWhile
\end{algorithmic}
\caption{Large-Context \algName{} Prediction \& Update}
\label{algo:lc_ow_pcoc}
\end{algorithm}

\subsubsection{Large-Context \algName{} (\algLargeContext)}
In the large-context setting, \algLargeContext{} operates in a very similar manner. The key practical difference between the settings is that we leverage the class embeddings, $(\bm{\mu}_e, \bm{\Sigma}_e)$, learned during pre-training. These embeddings are used to initialize class distributions for the $\numkk$ known-known classes during meta-training (discussed in the next subsection) forgoing the need to initialize these distributions from the shared prior at test-time. Indeed, the critical difference between the two settings is that in the LC setting, the information learning in pre-training and meta-training is known to be useful at test time. 

The shared prior continues to facilitate unknown-unknown class detection and unknown-known class instantiation. Initialization of the model in the large-context setting requires learned known-known class statistics $[\bm{q}_1,..., \bm{q}_\numkk]$ and $[\Lambda_1, ..., \Lambda_\numkk]$, shared prior statistics $(\bm{q}_0, \Lambda_0, \Sigma_\epsilon)$ and CRP statistics $(a, b)$. The algorithm then proceeds as outlined in the small-context setting with the exception that we do not update known-known class statistics ($y_* > \numkk$) because these distributions are learned directly. 

We initialize the class counts, $\bm{c}$, in the CRP as zero for all known-known classes. This may initially seem like an unusual design choice, as we clearly have observed data from the known-known classes. \algName{} assumes arbitrarily large-scale pre-training combined with a data-efficient online inference problem. We expect that the few-shot open-world recognition problem formulation is in fact motivated in some part by the existence of covariate shift between the large-scale training set and the data observed online. Thus, our model is technically misspecified as we do not explicitly consider this distributional shift between train and test, and moreover, algorithmic consideration of this requires explicit characterization of the shift and is thus challenging if not infeasible. We emphasize that these counts may be treated as a hyper-parameter, although we only consider the initialization at zero for all known-known classes. The procedure for this setting has been summarized in Algorithm \ref{algo:lc_ow_pcoc}.

\begin{algorithm}[t] 
\begin{algorithmic}[1]
    \Require{$\bm{\phi}, a, \Sigma_\epsilon, \alpha, \lambda,  \Dmetatrain$}
    \State Randomly initialize parameters: $b, \bm{q}_0, \Lambda_0$
    \While{not done}
        \State Sample Task: $\mathcal{T}_i = (\Dsupp, \Dquer) \sim \Dtrain$
        \State Initialize: $\Psi \leftarrow \text{\algName{}}(\bm{\phi}, \bm{q}_0, \Lambda_0, \Sigma_\epsilon, a, b, \Dsupp)$ \Comment{Alg. \ref{algo:sc-owpcoc} Phase 1}
        \State Let $\Dquery = (\bm{x}, \bm{y})$
        \State Predict: $\hat{\bm{y}} \leftarrow \Psi(\bm{x})$ \Comment{Eq. \ref{eq:prediction}}
        \State Evaluate: $\mathcal{L} \leftarrow \mathcal{L}_\text{NLL}(\hat{\bm{y}}, \bm{y}) + \lambda\mathcal{L}_\text{Adapt}(\Psi, \bm{x}, \bm{y})$
        \State Update: $(\bm{\phi}, b, \bm{q}_0, \Lambda_0) \leftarrow (\bm{\phi}, b, \bm{q}_0, \Lambda_0) - \alpha\nabla\mathcal{L}$
    \EndWhile
    \State \Return $(\bm{\phi}, \bm{q}_0, \Lambda_0, b)$
\end{algorithmic}
 \caption{Meta-Learning Small-Context \algName{}}
 \label{algo:ml-sc-owpcoc}
\end{algorithm}

\subsection{Meta-Learning \algName{}}
\label{sec:metalearning}

The flexibility of the \algName{} framework allows for meta-training in both the small and large-context settings with only minor changes to the training methodology. We assume access to a meta-training set, $\Dmetatrain$, consisting of $\Ntrain$ classes and sample meta-training tasks, $\mathcal{T}\sim\Dmetatrain$. These tasks correspond to a subset of the classes in the full dataset, and for each class, a subset of the data. In practice, we use the pre-training dataset $\Dtrain$ as the meta-training dataset. For classification purposes in the query phase of meta-training, there are $\numclasses+1$ classification labels: $\numclasses$ corresponding to the known classes and one corresponding to all unknown-unknown classes. Both settings initialize the encoder, $\bm{\phi}$, using weights from the pre-training phase. 

\subsubsection{Small-Context Meta-Learning}
Small-context meta-training tasks are constructed by sampling a support set $\Dsupp$ consisting of $\numclasses < \Ntrain$ classes and a query set $\Dquer$, such that $\Dsupp \cup \Dquer \subset \Dtrain$ and $\Dsupp \cap \Dquer = \emptyset$. To populate $\Dsupp$, we sample a small set of datapoints for all $\numclasses$ classes. We do not assume the number of images per support class is balanced. $\Dquer$ is generated by sampling a set of unknown-unknown classes from the $\Ntrain - \numclasses$ remaining classes. In the small-context setting, we learn the set of parameters $(\bm{\phi}, \bm{q}_0, \Lambda_0, b)$ given the the hyper-parameters $a, \Sigma_\epsilon$. \algSmallContext{} initializes, and computes the posterior class statistics given $\Dsupp$ (See Alg. \ref{algo:sc-owpcoc} Phase 1), without computing a loss. The model then evaluated over all datapoints in $\Dquer$, making predictions about whether images belong to a known or unknown-unknown class before computing a negative predictive log-likelihood (NLL) loss function, $\mathcal{L}_{\text{NLL}}$. See Algorithm \ref{algo:ml-sc-owpcoc} for the complete training procedure.

\subsubsection{Large-Context Meta-Learning}
Large-context meta-training tasks sample only a query set $\Dquer\sim\Dmetatrain$ consisting of data from $\numkk < \Ntrain$ known-known classes. We additionally sample a set of unknown-unknown classes from the remaining $\Ntrain - \numkk$ classes. We directly learn class means and covariances for each $\numkk$ class in addition to the set parameters $(\bm{\phi}, \bm{q}_0, \Lambda_0, b)$.  We initialize the additional parameters using the known-known class embeddings, $(\bm{\mu}_e, \bm{\Sigma}_e)$ learned during pre-training. The pre-training embeddings are factored using Equation \ref{eq:pcoc-factorization} to obtain the known-known class statistics $[\bm{q}_1, \dots, \bm{q}_\numkk]$ and $[\bm{\Lambda}_1, \dots, \bm{\Lambda}_\numkk]$. As in the small-context setting, the model evaluates over all datapoints in $\Dquer$, and computes a NLL loss function. The full training procedure is provided in Algorithm \ref{algo:ml-lc-owpcoc}.

\begin{algorithm}[t] 
\begin{algorithmic}[1]
    \Require{$\bm{\phi}, \alpha, \lambda, a, \bm{\mu}_e, \bm{\Sigma}_e, \Sigma_\epsilon, \Dmetatrain$}
    \State Randomly initialize: $b, \bm{q}_0, \Lambda_0$
    \State Factor: $[\bm{q}_1,..., \bm{q}_\numkk], [\Lambda_1, \dots, \Lambda_\numkk]  \leftarrow \bm{\mu}_e, \bm{\Sigma}_e$ \Comment{Eq. \ref{eq:pcoc-factorization}}
    \State Initialize: $\bm{q} \leftarrow [\bm{q}_0, \bm{q}_1, \dots, \bm{q}_\numkk]$
    \State Initialize: $\bm{\Lambda} \leftarrow [\Lambda_0, \Lambda_1, \dots, \Lambda_\numkk]$  
    
    \While{not done}
        \State Sample Task: $\mathcal{T}_i = (\Dquer) \sim \Dmetatrain$
        \State Initialize: $\Psi \leftarrow \text{\algName{}}(\bm{\phi}, \bm{q}_0, \Lambda_0, \Sigma_\epsilon, a, b, \numkk)$ \Comment{Alg. \ref{algo:lc_ow_pcoc} Phase 1}
        \State Let $\Dquery = (\bm{x}, \bm{y})$
        \State Predict: $\hat{\bm{y}} \leftarrow \Psi(\bm{x})$ \Comment{Eq. \ref{eq:prediction}}
        \State Evaluate: $\mathcal{L} \leftarrow \mathcal{L}_\text{NLL}(\hat{\bm{y}}, \bm{y}) + \lambda\mathcal{L}_\text{Adapt}(\Psi, \bm{x}, \bm{y})$
        \State Update: $(\bm{\phi}, b, \bm{q}, \bm{\Lambda}) \leftarrow (\bm{\phi}, b, \bm{q}, \bm{\Lambda}) - \alpha\nabla\mathcal{L}$
    \EndWhile
    \State \Return $(\bm{\phi}, \bm{q}, \bm{\Lambda}, b)$
\end{algorithmic}
 \caption{Meta-Learning Large-Context \algName{}}
 \label{algo:ml-lc-owpcoc}
\end{algorithm}

\subsubsection{Meta-Loss Function}
We note that the above is not strictly open-world training during the meta-training as there is no adaptation to novel classes. We find that better performance can be achieved by supervising the adaptation independently. We propose an adaptation loss, $\mathcal{L}_{\text{Adapt}}$,  which provides the model with direct supervision to its ability to adapt to unknown-known classes given a single labeled example. The complete loss function, $\mathcal{L}$ is therefore:

\begin{equation}
\label{eq:meta-loss}
    \mathcal{L} = \mathcal{L}_\text{NLL} + \lambda \mathcal{L}_{\text{Adapt}}
\end{equation}
where $\lambda$ is a loss weighting hyper-parameter. To evaluate the adaptation loss, a single datapoint, is sampled from each semantic class contained in the $\Dquer$ unknown-unknown class bucket. The posterior predictive distribution is computed for each class by independently conditioning the shared prior over embeddings on each respective sampled datapoint. The remaining data is classified via Bayesian Gaussian discriminant analysis and a negative log-likelihood loss (NLL) is computed. The adaptation loss evaluation procedure is provided in Algorithm \ref{algo:adapt}.

\subsection{Fine Tuning}
\label{sec:ft}

We fine-tune the final layer of the encoder in the small-context test phase to leverage the unknown-known class labels available in the support set. We employ the same procedure used during meta-training with the exception that we cannot compute the adaptation loss as, of course, no labels are available for query datapoints prior to evaluation. The loss is used to update the output layer---the last linear layer of the encoder generating the embeddings---via gradient descent. 

The fine-tuning of this linear output layer may initially seem surprising. Updating the last layer of the encoder corresponds to a different linear transformation of the nonlinear features, and thus redundant with the embedding adaptation. Because Gaussians are closed under linear transformations, this should not improve performance. The key detail is that while general (dense covariance) Gaussians are closed under linear maps, the set of isotropic Gaussians is not closed under linear transformation of the features. Thus, this fine-tuning step uses the small amount of provided support data at test time to re-scale the features to enable better modeling of the test data via isotropic Gaussians. 

In our experiments, we also investigated more aggressive fine-tuning schemes in which other layers of the encoder network were fine-tuned. We found that this was typically highly unstable. We also did not investigate combining in-episode fine-tuning with our approach. We view combining in-episode fine-tuning schemes, which potentially leverage ideas from continual learning \cite{kirkpatrick2017overcoming, ravichandran2020incremental, farajtabar2020orthogonal}, with our embedding-based open-world learning scheme as a particularly promising direction of future work beyond scope of this paper.

\begin{algorithm}[t] 
\begin{algorithmic}[1]
    \Require{$\bm{\phi}$, $\bm{q}_0, \Lambda_0, \Sigma_\epsilon, \Dquer$}
    \State Select unknown-unknown query class data: $\mathcal{D}_{apt} \leftarrow \Dquer(y = \numclasses+1)$
    \State Re-assign true class labels to $\mathcal{D}_{apt}$ classes.
    \State Initialize: $\bm{q} \leftarrow [\,]$ 
    \State Initialize: $\bm{\Lambda} \leftarrow [\,]$ 
    \For{n in $\mathcal{D}_{apt}$} 
        \State Sample datapoint: $(\bm{x}, y)\sim\mathcal{D}_{apt}$
        \State Append: $\bm{q} \leftarrow [\bm{q}, \bm{q}_0]$ 
        \State Append: $\bm{\Lambda} \leftarrow [\bm{\Lambda},  \Lambda_0]$
        \State Update $\bm{q}[y] \leftarrow \bm{q}[y] + \Sigma_\epsilon^{-1} \bm{\phi}(\bm{x})$
        \State Update $\bm{\Lambda}[y] \leftarrow \bm{\Lambda}[y] + \Sigma_\epsilon^{-1}$
    \EndFor
    \State Let $\mathcal{D}_{apt} = (\bm{x}, \bm{y})$
    \State Predict: $\hat{\bm{y}} = \text{Predict}(\bm{\phi}(\bm{x}), \bm{q}, \bm{\Lambda})$ \Comment{Eq. \ref{eq:prediction}} 
    \State Evaluate: $\mathcal{L} \leftarrow \mathcal{L}_\text{NLL}(\hat{\bm{y}}, \bm{y})$ 
    \State \Return $\mathcal{L}$
\end{algorithmic}
 \caption{Adaptation Loss}
 \label{algo:adapt}
\end{algorithm}

\subsection{Computational Complexity}
    We now discuss the test time computational complexity of the FLOWR framework. During prediction at time $t$ (equation \ref{eq:prediction}), we evaluate $N_t+1$ Gaussian distributions followed by a softmax operation. As discussed in section \ref{sec:computing-posterior-predictive} we restrict covariances to be isotropic, avoiding expensive matrix inversions which incur cubic complexity. Therefore, we can evaluate the distributions and the softmax in $\mathcal{O}(N_t \phidim)$ time where $N_t$ is the number of known classes and $\phidim$ is the dimension of the feature space. The prediction complexity dominates the recursive posterior parameter updates which run in $\mathcal{O}(\phidim)$ time. For vision tasks, the overall runtime of the approach is dominated by the computational cost of the feature extractor. For example, evaluating a forward pass of a Resnet18 requires $1.8$ billion FLOPS (multiply-adds) \cite{he2016deep} which far exceeds the cost of the FLOWR framework. This highlights a strength of our approach: adaptation and instantiation of novel classes corresponds solely to operations on the output layer of the neural network model. Thus, at time $t$, the neural network backbone (or encoder) must only be evaluated once, substantially reducing test time complexity relative to other meta-learning-based models.

\section{Experiments}
\label{sec:experiments}

We evaluate our proposed method using the MiniImageNet \cite{vinyals2016matching} and TieredImageNet \cite{ren2018meta} datasets. Both datasets are subsets of ImageNet ILSVRC-12 \cite{imagenet_cvpr09}. MiniImageNet contains 100 classes, each with 600 images. We implement the MiniImageNet class splits proposed by \cite{ravi2017optimization}, which segment the dataset into 64 training, 12 validation and 24 test classes. TieredImageNet is significantly larger, containing 608 classes and over 700,000 images. The ImageNet hierarchical structure is leveraged to construct meta-train and meta-test class splits which minimize the semantic similarity of classes across the split. The result is a more challenging and realistic generalization task. In the large-context setting we use the additional data provided by \cite{gidaris2018dynamic} for evaluating the MiniImageNet meta-train split classes. TieredImageNet meta-train split classes are evaluated by reserving approximately 20\% of the meta-train class data for validation and testing. We use the same partitions proposed by~\cite{ren2019attractor}. All images, across both datasets, are resized to a uniform 84x84 dimension. 

We utilize two encoder architectures in our experiments: the Conv-4 network architecture proposed by \cite{vinyals2016matching} for MiniImageNet experiments and a larger Resnet18 network architecture \cite{he2016deep} for the TieredImageNet experiments. The Conv-4 architecture consists of 4 stacked network modules.  Each module contains a 3x3 convolutional layer with 64 filters, batch normalization, a ReLU activation and a 2x2 max pooling layer. Finally we augment each network with an additional linear layer which reduces the dimensionality of each respective encoder feature space to $64$.The CRP hyper-parameter, $a$, is set to be $0.5$ in all experiments. $\Sigma_\epsilon$ is set as $0.5$. We use learning rates of $1e^{-3}$ and $1e^{-4}$ for the Conv4 and Resnet networks respectively. We set both loss weighting parameters , $\beta$ and $\lambda$ to be $0.1$.




\rev{\begin{table}[]
    \caption{PEELER AUROC vs. H-Measure}
    \label{tab:auroc_demo}
    \centering
\begin{tabular}{cccc}
\hline
\textbf{Method} & \textbf{AUROC (\%)} & \textbf{H - B(2, 2) (\%) } & \textbf{H - B(2, 1)(\%)} \\ \hline
EVM & 53.24 & 1.00 & 0.14\\
PEELER & 54.96 &  2.36 & 0.15
\end{tabular}
\end{table}}

\begin{figure}[t]
\begin{center}
\includegraphics[width=\linewidth]{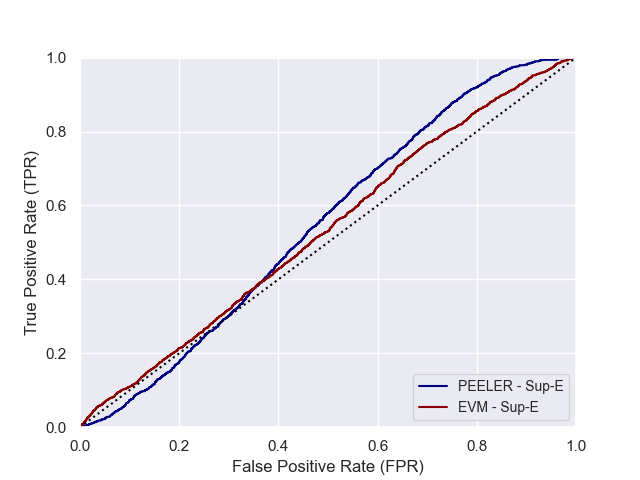}

\end{center}
   \captionsetup{width=.8\linewidth}
   \caption{PEELER/EVM novel class detection ROC curves for Small-Context TieredImageNet experiments. }
\label{fig:auroc_demo}
\end{figure}

\begin{table*}[]
    \caption{TieredImageNet - Small-Context Few-Shot Open-World Recognition Results}
    \label{tab:tiered-sc-results}
    \centering
\begin{tabular}{cccccc}
\hline
\textbf{Method} & \textbf{Pre-Training} & \textbf{Acc. (\%)} & \textbf{Support Acc. (\%)} & \textbf{Inc. Acc. (\%)} & \textbf{H-Measure (\%)} \\ \hline
NCM             & Sup-FC                & 45.71$\pm$0.71         & 51.62$\pm$0.09                 & 33.88$\pm$0.46              & 0.18$\pm$0.10                \\
NCM             & Sup-E                 & 44.09$\pm$0.28         & 49.32$\pm$0.27                 & 33.65$\pm$0.61              & 0.41$\pm$0.26               \\
ProtoNet        & Sup-FC                & 46.71$\pm$0.25         & 54.16$\pm$0.22                 & 31.62$\pm$0.45              & 0.54$\pm$0.15               \\
ProtoNet        & Sup-E                 & 48.52$\pm$0.22         & 54.94$\pm$0.31                 & 35.67$\pm$0.48              & 0.49$\pm$0.14               \\
PEELER          & Sup-FC                & 38.28$\pm$0.72              & 42.98$\pm$0.36                      & 28.90$\pm$0.23                   & 1.77$\pm$0.13                    \\
PEELER          & Sup-E                 & 39.09$\pm$0.68              & 44.76$\pm$0.39                      & 27.76$\pm$0.42                   & 2.36$\pm$0.20                    \\
EVM             & Sup-FC                & 40.08$\pm$0.33              & 47.53$\pm$0.27                    & 27.48$\pm$0.36                   & 1.62$\pm$0.17                   \\
EVM             & Sup-E                 & 41.51$\pm$0.29              & 48.38$\pm$0.35                      & 27.76$\pm$0.22                   & 1.00$\pm$0.35                    \\
\algSmallContext{} (Ours)  & Sup-FC                & 49.59$\pm$0.15         & 55.63$\pm$0.28                 & 37.50$\pm$0.51              & 13.31$\pm$0.45              \\
\algSmallContext{} (Ours)  & Sup-E                 & $\bm{51.64}\pm0.14$         & $\bm{57.76}\pm0.20$                 & $\bm{39.39}\pm0.04$              & $\bm{15.04}\pm0.77$              \\ \hline
\end{tabular}
\end{table*}

\begin{table*}[]
  \caption{MiniImageNet - Small-Context Few-Shot Open-World Recognition Results}
  \label{tab:mini-sc-results}
  \centering
\begin{tabular}{cccccc}
\hline
\textbf{Method} & \textbf{Pre-Training} & \textbf{Acc. (\%)} & \textbf{Support Acc. (\%)} & \textbf{Inc. Acc. (\%)} & \textbf{H-Measure (\%)} \\ \hline
NCM             & Sup-FC                & 37.88$\pm$0.91         & 43.45$\pm$0.94                 & 26.75$\pm$0.22              & 1.78$\pm$0.27               \\
NCM             & Sup-E                 & 27.49$\pm$0.08         & 32.07$\pm$0.07                 & 18.32$\pm$0.09              & 8.12$\pm$0.40                \\
ProtoNet        & Sup-FC                & 30.38$\pm$0.13         & 35.31$\pm$0.13                 & 20.51$\pm$0.12              & 2.39$\pm$0.28               \\
ProtoNet        & Sup-E                 & 34.85$\pm$0.12         & 40.55$\pm$0.16                 & 23.45$\pm$0.14              & 1.62$\pm$0.15               \\
PEELER          & Sup-FC                & 36.12$\pm$0.22         & 38.32$\pm$0.26                       & $31.71\pm0.19$                    & 2.05$\pm$0.09                    \\
PEELER          & Sup-E                 & 37.40$\pm$0.17              & 39.04$\pm$0.15                      & $\bm{34.09}\pm$0.17                   & 1.11$\pm$0.11                    \\
EVM          & Sup-FC                & 32.12$\pm$0.56              & 32.72$\pm$0.57                      & 14.73$\pm$0.32                   & 0.61$\pm$0.32                    \\
EVM          & Sup-E                 & 30.26$\pm$0.32             & 30.71$\pm$0.31                      & 17.4$\pm$0.24                   & 0.34$\pm$0.12                    \\
\algSmallContext{} (Ours)  & Sup-FC                & 38.86$\pm$0.07         & 46.10$\pm$0.10                 & 24.39$\pm$0.13              & 8.64$\pm$0.38               \\
\algSmallContext{} (Ours)  & Sup-E                 & $\bm{41.27}\pm0.10$         & $\bm{51.88}\pm0.18$                 & 20.05$\pm$0.15              & $\bm{19.06}\pm0.88$              \\ \hline
\end{tabular}
\end{table*}

\subsection{Evaluation Methodology} 

The multi-task nature of open-world recognition makes it difficult to define a single performance metric. Classification performance has typically been evaluated via top-1 classification accuracy~\cite{rudd2018evm} while the area under the receiver operating characteristic (AUROC) is also often reported as a metric for unknown-unknown class detection~\cite{liu2020openset, geng2020recent}. 

The use of AUROC in particular is problematic because it is not a model-independent metric. Hand~\cite{hand2009hmeasure} demonstrates that computing AUROC implicitly defines distributions over misclassification cost weighting which are a function of the empirical scoring distributions of the models themselves. Practically this is equivalent to utilizing a different performance metric to evaluate each classifier, rendering comparisons between classifiers based on AUROC invalid. Hand proposes the H-Measure as an alternative to AUROC which fixes a Beta(2, 2) distribution over misclassification costs. While Hand acknowledges that this is an arbitrary choice of distribution and that the cost of misclassification is domain specific, this choice is deterministic and decouples the model from the performance metric and provides a sound basis for comparison. Compared to the uniform distribution, another simple choice, the beta distribution has the advantage of decaying towards zero for extreme values of misclassification cost. Note an AUROC score of $50\%$ and an H-Measure score of $0\%$ correspond to random chance (diagonal ROC curve) while a score of $100\%$ corresponds to perfect classification for both metrics.

We isolate unknown-unknown class detection results from the PEELER and EVM baseline to demonstrate the unreliability of AUROC. We compare these baselines in the small-context setting. The ROC curves of both methods have been superimposed in Figure \ref{fig:auroc_demo}. When comparing the curves, we see that their classification performance differs significantly depending on the operating point, however, both methods report similar AUROC scores. In Table \ref{tab:auroc_demo}, we see that EVM reports a score of $53.24$ and PEELER a score of $54.96$. We expect that the difference between the methods should be reflected by our evaluation metric. The H-measure provides a principled solution through explicit specification of a distribution over misclassification cost. Under the default Beta(2, 2) distribution, EVM and PEELER report H-measure scores of $1.00$ and $2.36$ respectively. If we instead impose a Beta(2, 1) distribution over misclassification cost, which increases the weighting on higher costs of misclassification, we increase the impact of performance at lower true positive rates. We can see this impact reflected with H-measure scores of $0.14$ and $0.15$ for EVM and PEELER respectively.


Some methods \cite{neal2018open, liu2020openset} also choose to report a closed-set accuracy metric alongside AUROC. Reporting the closed-set accuracy is problematic because the known class accuracy is fundamentally linked to the task of unknown-unknown class detection. The link is intuitive when considering thresholded methods that must explicitly select an operating point (threshold). The threshold cannot be ignored when evaluating accuracy while simultaneously reporting an integral metric for unknown-unknown detection. For the same reason, we must also fix the operating point when directly comparing methods to facilitate a fair comparison of accuracy metrics.

To alleviate these concerns, we report the the overall top-1 classification accuracy for each method, and further decompose the results into support and incremental classification accuracy, which capture the models' ability to classify known classes and to rapidly adapt to unknown-unknown classes once a label is provided. We fix the unknown-unknown detection true positive rate (TPR) across all compared models to report the accuracy metrics at a consistent operating point for each experiment. The performance metrics are summarized as follows:

\begin{itemize}
    \item \textbf{Accuracy}: Total classification accuracy of the model with respect to all data observed during evaluation. This metric captures the classification ability of the model with respect to all known-known, unknown-known and unknown-unknown class data.
    \item \textbf{Support-Accuracy}: Support accuracy computes the classification accuracy of the model with respect to classes (either known-known or unknown-known) observed by the model prior to evaluation.
    \item \textbf{Incremental-Accuracy}: Incremental accuracy computes the accuracy of the model with respect to all unknown-unknown class data (unobserved prior to evaluation). This metric captures the ability of the model to quickly adapt to novel classes.
    \item \textbf{Novel Class H-Measure}: The novel-class H-Measure provides a measure of the model's ability to discriminate between unknown-unknown and known classes (either known-known or unknown-known). It is important to note that once an unknown-unknown class label has been received, all subsequent data of that class is considered to belong to an unknown-known class.
\end{itemize}

\subsection{Small-Context Results}
\label{subsec:small-context}

\textbf{Task Sampling.} Small-Context FS-OWR meta-training tasks are sampled from each dataset's meta-train class split. During training, the support set, $\Dsupp$, samples $40$ classes. An additional $10$ classes are sampled for inclusion in the query set, $\Dquery$, to represent unknown-unknown classes. Small-Context FS-OWR test tasks sample $10$ support classes from the meta-test class splits and an additional $5$ classes for the query set. Each support set class samples between $[1,10]$ datapoints and all classes sample 10 datapoints to build the query set. At both meta-training and meta-test time, the task query sequence is generated by randomly permuting the query set. Each model is trained for 60 epochs (each consisting of 1000 tasks) and is evaluated across 1000 tasks.

\noindent\textbf{Baselines.}
The performance of \algSmallContext{} is benchmarked with respect to Nearest Class Mean, ProtoNet, PEELER and EVM baselines. In each case we compare our supervised-embedding pre-training (Sup-E) with fully connected classification head pre-training (Sup-FC), as in \cite{chen2020new}. Pre-training is performed using the meta-train class splits.
\begin{itemize}
    \item \textbf{Nearest class mean (NCM)} has shown strong baseline performance in the closed-world few-shot setting \citep{wallingford2020wild, chen2020new}. Class-wise means of support feature vectors are computed and then NCM performs top-1 nearest-neighbors classification via Euclidean distance. We adapt the NCM baseline for use in the open-world setting by thresholding the top-1 classification with a tunable minimum distance. The encoder weights are trained non-episodically on the meta-train splits using standard supervised learning. At test time, the class means are updated after each observed label. 
    \item \textbf{Prototypical networks (ProtoNet)} follow the implementation and training protocol proposed by \cite{snell2017prototypical}. It is similarly adapted for novel class detection via a tuned threshold and the class prototypes are also updated online after each observed label.
    \item \textbf{PEELER} (oPen sEt mEta LEaRning) \cite{liu2020openset} is a few-shot open-set recognition algorithm which maintains a Gaussian embedding for each known class and performs nearest-neighbors classification via Mahalanobis distance. Class-wise means of support feature vectors are used to initialize distribution means while a fully-connected head estimates a precision matrix from each class mean. Novel class detection is performed via thresholding the normalized Mahalanobis distances. 
    \item \textbf{Extreme Value Machine (EVM)}\cite{rudd2018evm} is a prominent OWR classifier. It is non-differentiable and relies on a pre-trained encoder, therefore, we utilize the same encoder weights as the NCM baseline. We use cosine-distance for computing margin-distances (as recommended by \cite{rudd2018evm}) and a tailsize of 10. 
\end{itemize}

\begin{table*}[]
    \caption{TieredImageNet - Large-Context Few-Shot Open-World Recognition Results}
    \label{tab:tiered-lc-results}
    \centering
\begin{tabular}{ccccc}
\hline
\textbf{Method} & \textbf{Acc. (\%)} & \textbf{Support Acc. (\%)} & \textbf{Inc. Acc. (\%)} & \textbf{H-Measure (\%)} \\ \hline
NCM & $25.95\pm0.38$ & $26.07\pm0.24$ & $20.84\pm0.33$ & $1.41\pm0.20$ \\
ProtoNet & $28.31\pm0.28$ & $28.64\pm0.25$ & $14.93\pm0.30$ & $2.27\pm0.16$ \\
EVM & $16.02\pm0.75$ & $18.51\pm0.68$ & $14.89\pm0.34$ & $0.79\pm0.19$ \\
ACastle & $\mathbf{33.49}\pm0.27$ & $\mathbf{34.00}\pm0.25$ & $19.54\pm0.41$ & $8.49\pm0.34$ \\
\algLargeContext{} (Ours)  & $24.93\pm0.17$ & $24.36\pm0.15$ & $\mathbf{47.66}\pm0.45$ & $\mathbf{14.58}\pm0.61$ \\ \hline
\end{tabular}
\end{table*}

\begin{table*}[]
    \caption{MiniImageNet - Large-Context Few-Shot Open-World Recognition Results }
    \label{tab:mini-lc-results}
    \centering
\begin{tabular}{ccccc}
\hline
\textbf{Method} & \textbf{Acc. (\%)} & \textbf{Support Acc. (\%)} & \textbf{Inc. Acc. (\%)} & \textbf{H-Measure (\%)} \\ \hline
NCM             & $22.71\pm0.55$              & $23.13\pm0.52$                      & $17.31\pm0.66$                   & $0.98\pm0.18$                       \\
ProtoNet        & $\mathbf{28.71}\pm0.31$   & $\mathbf{30.22}\pm0.35$                      & $9.35\pm0.62$                    & $2.65\pm0.37$                       \\
EVM             & $15.44\pm0.82$              & $16.19\pm0.76$                      & $5.9\pm0.74$                     & $0.90\pm0.33$                    \\
AA-Network      & $26.06\pm0.26$              & $25.06\pm0.23$                      & $39.12\pm0.29$                   & $14.37\pm0.50$                       \\
ACastle      & $26.43\pm0.33$              & $27.99\pm0.29$                      & $12.12\pm0.48$                   & $3.45\pm0.36$                       \\
\algLargeContext{} (Ours)  & $21.64\pm0.20$              & $19.33\pm0.22$                      & $\mathbf{51.15}\pm0.31$                   & $\mathbf{25.41}\pm0.86$                   \\ \hline
\end{tabular}
\end{table*}

\begin{figure*}
\centering
    \begin{subfigure}{0.48\textwidth}
        \includegraphics[width=\linewidth]{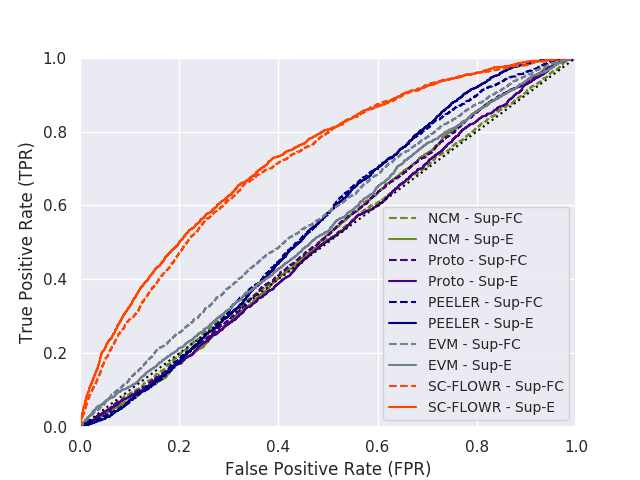}
        \caption{Small Context - TieredImageNet}
        \label{fig:roc_a}
    \end{subfigure}
    \begin{subfigure}{0.48\textwidth}
        \includegraphics[width=\linewidth]{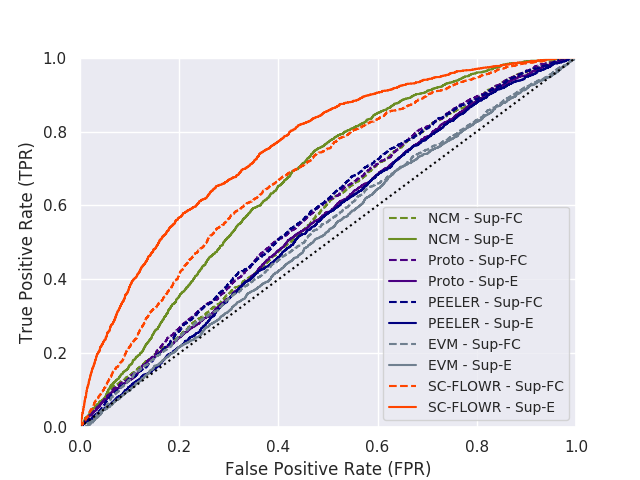}
        \caption{Small Context - MiniImageNet}
        \label{fig:roc_b}
    \end{subfigure}
    \begin{subfigure}{0.48\textwidth}
        \includegraphics[width=\linewidth]{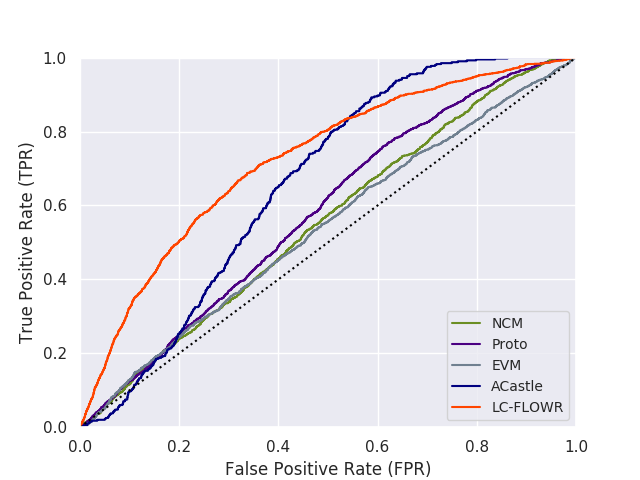}
        \caption{Large Context - TieredImageNet}
        \label{fig:roc_c}
    \end{subfigure}
    \begin{subfigure}{0.48\textwidth}
        \includegraphics[width=\linewidth]{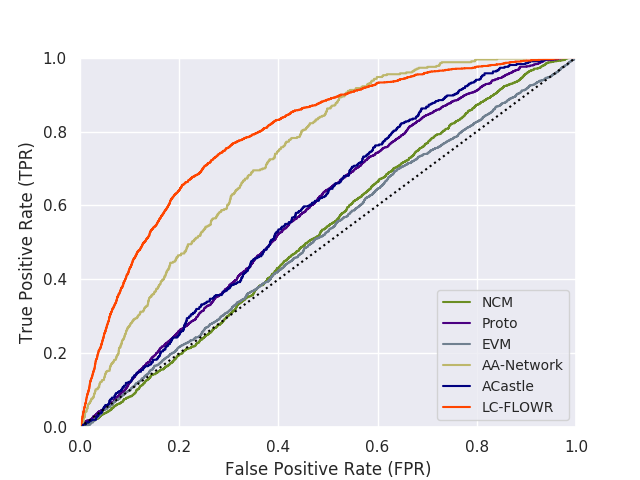}
        \caption{Large Context - MiniImageNet}
        \label{fig:roc_d}
    \end{subfigure}
\caption{Novel Class Detection ROC Curves}
\label{fig:lc-lpcoc-roc}
\end{figure*}

\noindent\textbf{Results.}
We present the small-context results on the MiniImageNet and TieredImageNet datasets in Tables \ref{tab:tiered-sc-results} and \ref{tab:mini-sc-results}. Accuracy metrics are reported at a unknown-unknown detection TPR of 0.15 and ROC curves for each method are provided in Figures \ref{fig:roc_a} and \ref{fig:roc_b}. 
We observe an improvement in classification performance when initializing the encoder weights with our supervised embedding pre-training scheme for all of the methods except NCM. This suggests that a generative-modeling assumption during pre-training learns features that improve generalization in the open-world setting regardless of the closed-world assumption during pre-training. On the MiniImageNet dataset, the NCM classification accuracy also outperformed meta-trained algorithms. This result has been previously observed by \cite{chen2020new} and \cite{triantafillou2019meta} who both demonstrate that meta-learning may be harmful to generalization in some visual settings. 

\algSmallContext{} demonstrates strong classification performance on both datasets. On the larger-scale, more challenging TieredImageNet dataset, \algSmallContext{} outperforms the baselines in total accuracy, support and incremental accuracy. An important feature of the \algSmallContext{} algorithm is that it allows for end-to-end training in the small-context FSOWR setting. Unlike the baselines, \algSmallContext{} does not depend on non-differentiable thresholding to detect novel classes. The model can therefore learn to calibrate predictions given the significant imbalance between novel class and known class data. \algSmallContext{} significantly outperforms the baselines when detecting novel classes on both datasets (as measured by H-Measure). The NCM and Protonet baselines do not demonstrate a capacity to detect novel classes on either dataset. The PEELER and EVM baselines show some ability to discriminate, however, they struggle to adapt to this setting. The EVM baseline in particular may struggle because the underlying assumptions of extreme value theory rely on the assumption of a large dataset and do not necessarily hold in a few-shot setting. 

From our experiments, we observe that the embedding-based pre-training improves classification accuracy and \algSmallContext{} results in substantially improved H-Measure without degrading classification accuracy. Thus, the combination of embedding-based pre-training and \algSmallContext{} results in a robust model with balanced performance across all metrics necessary for deployment in the small-context few-shot open-world recognition setting.

\subsection{Large-Context Results} 

In the large-context setting, we restrict our attention to the Sup-E pre-training approach, based on the strong results across all methods from Section~\ref{subsec:small-context}. 

\noindent\textbf{Task Sampling.} Large-Context FS-OWR tasks sample $5$ unknown-unknown classes in addition to the known-known classes carried forward from meta-training or pre-training time. Each task samples $10$ query datapoints from each class to build the query set. The task test sequence is generated by randomly permuting the query set. MiniImageNet tasks contain $64$ known-known classes and TieredImageNet tasks contain of $200$ known-known classes. We note that in comparison to the small-context setting, a significantly more challenging $64$+$5$-way or $200$+$5$-way classification task is performed. Each model is trained for 60 epochs and evaluated over 1000 tasks.

\noindent\textbf{Baselines.} The baselines for the large-context setting have been adapted similarly to those in the small-context setting. For both the NCM and ProtoNet baselines, the train set is conditioned-on offline to generate class means for each known-known class. The non-parameteric nature of the EVM algorithm as implemented by \cite{rudd2018evm} does not scale in large-context setting resulting in an approximate run-time of 2 weeks for evaluation to complete compared to a matter of minutes for the other algorithms. We significantly sub-sample known-known class data to 50 datapoints per class in order to achieve a reasonable run-time. Similarly, PEELER has been omitted from the large-context setting because the large-scale variant of the algorithm proposed by \cite{liu2020openset} is too computationally expensive, requiring the model to re-train from scratch each time an additional class is incorporated.
\begin{itemize}
    \item \textbf{Attention-Attractor Networks (AA-Networks)} \cite{ren2019attractor}  is a few-shot incremental learning algorithm which combines a base learner for known-known classes with a few-shot learner for unknown-known classes. We adapt it for the large-context setting by thresholding class probabilities and re-training the few-shot learner as new labels are provided. The open-source implementation provided by \cite{ren2019attractor} does not include code necessary for training with TieredImageNet and re-implementation is challenging, and therefore we only provide baseline results on MiniImageNet. 
    \item \textbf{Adaptive Classifier Synthesis Learning (ACastle)} \cite{ye2021learning} is similar to AA-networks in architecture, combining a base learner for known-known classes with a few-shot learner for unknown-known classes. The joint classifier is synthesized using the base-learner and few-shot learner as query to a learnable neural dictionary. Class probabilities are thresholded to detect novel classes and the few-shot learner is updated as new labels are available.
\end{itemize}
 
\noindent\textbf{Results.} We present the large-context results in Table \ref{tab:tiered-lc-results} and associated ROC curves in Figures \ref{fig:roc_a} and \ref{fig:roc_b}. Accuracy metrics are reported at a unknown-unknown detection TPR of 0.6. 

H-measure performance overall remains low, which we attribute this to the difficulty incurred by the imbalance between novel class true positives and true negatives. We observe that there is a significant H-Measure increase in the baseline algorithms, when compared to the small-context setting. The baselines take advantage of the additional context available to form better open-space decision boundaries. The EVM baseline results are particularly poor in this setting, while the classification accuracy remains somewhat competitive, the sub-sampling has deteriorated its ability to detect unknown-unknown class data. 

The overall classification accuracy of \algLargeContext{} is competitive with baseline methods while demonstrating a significant improvement in both incremental accuracy and H-Measure. The combination of the known-known class embedding initialization and shared prior is conceptually similar to the base learner and few-shot learner pair proposed by both AA-Networks and ACastle. The strong performance of these methods compared to other baselines suggests that this is a promising approach. We also note that \algLargeContext{}, AA-Networks and ACastle also report strong H-measure results in comparison to the other methods. We attribute these results to their ability to dynamically model the variance of the class distributions. \algLargeContext{} in particular provides principled means of independently updating the decision boundary of each class through the covariance update. NCM and ProtoNet, in comparision, are less flexible models, given that they only update the class mean. When only a few examples are available per class, or when there is class imbalance the capability to model the variance of the class distributions is very helpful for novel class detection.


\begin{table}[]
  \caption{\algSmallContext{} Ablation - TieredImageNet}
  \label{tab:sc-ablation}
  \centering
 \resizebox{\columnwidth}{!}{  
\begin{tabular}{cccccc}
\hline
\textbf{AL} & \textbf{FT} & \textbf{Acc.} & \textbf{Sup. Acc.} & \textbf{Inc. Acc.} & \textbf{H-Measure} \\ \hline
    -              &          -            & $46.24\pm0.64$                 & $51.21\pm0.83$                         & $36.24\pm0.43$                       & $11.94\pm0.55$                   \\
\checkmark                 &          -            & $48.16\pm0.13$              & $53.74\pm0.11$                          & $37.02\pm0.13$                       & $13.29\pm0.59$                   \\
\checkmark                 & \checkmark                    & $\mathbf{51.64}\pm0.14$         & $\mathbf{57.76}\pm0.20$                 & $\mathbf{39.39}\pm0.04$              & $\mathbf{15.04}\pm0.77$                   \\
\end{tabular}}
\end{table}

\begin{table}[]
  \caption{CRP Ablation - TieredImageNet}
  \label{tab:crp-ablation}
  \centering
 \resizebox{\columnwidth}{!}{  
\begin{tabular}{cccccc}
\hline
\textbf{CRP}  & \textbf{Acc.} & \textbf{Sup. Acc.} & \textbf{Inc. Acc.} & \textbf{H-Measure} \\ \hline
    -              & $\mathbf{52.36}\pm0.11$                 & $\mathbf{57.58}\pm0.16$                         & $\mathbf{41.92}\pm0.07$                       & $13.54\pm0.93$                   \\
\checkmark         & $51.64\pm0.14$         & $\mathbf{57.76}\pm0.20$                 & $39.39\pm0.04$              & $\mathbf{15.04}\pm0.77$                   \\
\end{tabular}}
\end{table}

\begin{table}[!]
  \caption{Trace Regularization Ablation - MiniImageNet}
  \label{tab:beta-ablation}
  \centering
\begin{tabular}{ccc}
\hline
\textbf{Trace Reg.} & \textbf{Beta}($\beta$)& \textbf{Accuracy(\%)} \\ \hline
 - & - & 42.26 \\
\checkmark & 0.01 & 42.34 \\
\checkmark & 0.1 & $\mathbf{44.44}$ \\
\checkmark & 1.0 & 42.73 \\
\end{tabular}
\end{table}

\subsection{Ablation Studies}
We provide an ablation of the \algSmallContext{} training augmentations in Table \ref{tab:sc-ablation} and an ablation of the Chinese Restaurant Process in Table \ref{tab:crp-ablation}. The ablations are performed on TieredImageNet using a Resnet18 encoder. In this table, AL indicates the addition of the \textit{adaptation loss} and FT indicates the addition of \textit{fine-tuning}.

We specifically ablate the adaptation loss and test-time fine-tuning with respect to a baseline \algSmallContext{} model. When compared to the baseline, the adaptation loss increases the incremental accuracy as expected, however, it also improves the ability of the model to classify support class data. The fine-tuning step also provides a significant improvement to the \algSmallContext{} classification accuracy across all three metrics. Additionally, we observe an increase in the ability to discriminate novel classes. 

We compare the selection of the CRP with a uniform distribution over classes (the same assumption made during pre-training). We find that the learned CRP increases the capability of the model to identify unknown-unknown data, however, it comes at the cost of decreased incremental accuracy. We note that the selection of the CRP prior violates the assumption made in Section \ref{sec:problem} that class data is sampled i.i.d. The CRP rather assumes that the ''rich get richer`` where query data is more likely to belong to classes with more previous observations. This is a useful property when operating in environments with large class imbalances. However, following the procedure of \cite{triantafillou2019meta}, we do not evaluate under this sampling assumption because we seek equal performance across all classes . Ultimately, the selection of the class prior provides a useful mechanism to tune the performance of the algorithm.

We evaluate the effect of the trace regularization during pre-training in Table \ref{tab:beta-ablation}. The ablation is performed using the Conv4 backbone and the MiniImageNet dataset. We baseline the Sup-E pre-training without the regularization and compare a range of values for the regularization strength, $\beta$. We report the $64$-way classification accuracy on the MiniImageNet validation set proposed by~\cite{ren2019attractor}. With appropriate tuning of $\beta$, we observe a 2.18\% increase in classification accuracy.

\section{Conclusions}
\label{sec:conclusion}

In this work, we motivate the need to reformulate learning systems from a closed-world setting to an open-world, few-shot setting. We employ H-measure as a new metric performance metric for unknown-unknown class detection in these settings and demonstrate the unreliability of AUROC for fair model comparison. We present the \algNameLong{} (\algName{}) framework which combines Bayesian embedding-based meta-learning with a Chinese restaurant process class prior and a supervised embedding pre-training scheme which can be flexibly applied to both the small- and large-context settings. The framework outperforms baselines on MiniImageNet and TieredImageNet and demonstrates significant capability of detecting and quickly learning novel classes given few labels. 

There are two promising directions of future work for the framework developed in this paper. First, we do not update known-known class embeddings at test time in the large context setting, as doing so in the current implementation degrades classification performance. Nevertheless, the test time labels are valuable information that can be leveraged in future work to improve open-world performance and robustness to known-known class concept drift. Second, we have assumed throughout this paper that we are operating in a fully supervised setting in which labels are provided immediately after classification decisions are provided by the learner. A more realistic setting is a semi-supervised one in which labels are not always provided. While we could simply ignore unlabeled data in our class updates, this is ignoring a substantial amount of information available in observed covariate shifts. Thus, incorporating this unsupervised data in a principled fashion is an important avenue of future work. 

\appendices

\ifCLASSOPTIONcompsoc
  \section*{Acknowledgments}
\else
  \section*{Acknowledgment}
\fi

This research was supported in part by the National Aeronautics and Space Administration (NASA) via an Early Stage Innovations grant. James Harrison was also supported in part by the Natural Sciences and Engineering Research Council of Canada (NSERC) and Stanford University via the Stanford Graduate Fellowship.

\ifCLASSOPTIONcaptionsoff
  \newpage
\fi

{\small
\bibliography{cites}

\begin{thebibliography}{10}\itemsep=-1pt

\bibitem{allen2019infinite}
Kelsey~R Allen, Evan Shelhamer, Hanul Shin, and Joshua~B Tenenbaum.
\newblock Infinite mixture prototypes for few-shot learning.
\newblock {\em International Conference on Machine Learning (ICML)}, 2019.

\bibitem{bendale2015openworld}
Abhijit Bendale and Terrance Boult.
\newblock Towards open world recognition.
\newblock {\em {IEEE} Conference on Computer Vision and Pattern Recognition
  (CVPR)}, 2015.

\bibitem{bendale2016towards}
Abhijit Bendale and Terrance~E Boult.
\newblock Towards open set deep networks.
\newblock 2016.

\bibitem{bertinetto2018meta}
Luca Bertinetto, Jo{\~a}o~F Henriques, Philip~HS Torr, and Andrea Vedaldi.
\newblock Meta-learning with differentiable closed-form solvers.
\newblock {\em arXiv:1805.08136}, 2018.

\bibitem{castro2018end}
Francisco~M Castro, Manuel~J Mar{\'\i}n-Jim{\'e}nez, Nicol{\'a}s Guil, Cordelia
  Schmid, and Karteek Alahari.
\newblock End-to-end incremental learning.
\newblock {\em European Conference on Computer Vision (ECCV)}, 2018.

\bibitem{chen2020new}
Yinbo Chen, Xiaolong Wang, Zhuang Liu, Huijuan Xu, and Trevor Darrell.
\newblock A new meta-baseline for few-shot learning.
\newblock {\em arXiv:2003.04390}, 2020.

\bibitem{imagenet_cvpr09}
J. Deng, W. Dong, R. Socher, L.-J. Li, K. Li, and L. Fei-Fei.
\newblock {ImageNet: A Large-Scale Hierarchical Image Database}.
\newblock {\em {IEEE} Conference on Computer Vision and Pattern Recognition
  (CVPR)}, 2009.

\bibitem{farajtabar2020orthogonal}
Mehrdad Farajtabar, Navid Azizan, Alex Mott, and Ang Li.
\newblock Orthogonal gradient descent for continual learning.
\newblock {\em Artificial Intelligence and Statistics (AISTATS)}, 2020.

\bibitem{finn2017model}
Chelsea Finn, Pieter Abbeel, and Sergey Levine.
\newblock Model-agnostic meta-learning for fast adaptation of deep networks.
\newblock {\em International Conference on Machine Learning (ICML)}, 2017.

\bibitem{fisher1928limiting}
Ronald~Aylmer Fisher and Leonard Henry~Caleb Tippett.
\newblock Limiting forms of the frequency distribution of the largest or
  smallest member of a sample.
\newblock {\em Mathematical Proceedings of the Cambridge Philosophical
  Society}, 1928.

\bibitem{fox2009bayesian}
Emily~Beth Fox.
\newblock {\em Bayesian nonparametric learning of complex dynamical phenomena}.
\newblock PhD thesis, Massachusetts Institute of Technology, 2009.

\bibitem{geng2020recent}
Chuanxing Geng, Sheng-jun Huang, and Songcan Chen.
\newblock Recent advances in open set recognition: A survey.
\newblock {\em IEEE Transactions on Pattern Analysis \& Machine Intelligence},
  2020.

\bibitem{gershman2012tutorial}
Samuel~J Gershman and David~M Blei.
\newblock A tutorial on bayesian nonparametric models.
\newblock {\em Journal of Mathematical Psychology}, 2012.

\bibitem{gidaris2018dynamic}
Spyros Gidaris and Nikos Komodakis.
\newblock Dynamic few-shot visual learning without forgetting.
\newblock 2018.

\bibitem{gordon2018meta}
Jonathan Gordon, John Bronskill, Matthias Bauer, Sebastian Nowozin, and
  Richard~E Turner.
\newblock Meta-learning probabilistic inference for prediction.
\newblock {\em International Conference on Learning Representations (ICLR)},
  2019.

\bibitem{grant2018recasting}
Erin Grant, Chelsea Finn, Sergey Levine, Trevor Darrell, and Thomas Griffiths.
\newblock Recasting gradient-based meta-learning as hierarchical {B}ayes.
\newblock {\em International Conference on Learning Representations (ICLR)},
  2018.

\bibitem{hand2009hmeasure}
David Hand.
\newblock Measuring classifier performance: a coherent alternative to the area
  under the roc curve.
\newblock {\em Machine learning}, 2009.

\bibitem{harrison2021uncertainty}
James Harrison.
\newblock {\em Uncertainty and Efficiency in Adaptive Robot Learning and
  Control}.
\newblock PhD thesis, Stanford University, 2021.

\bibitem{harrison2019continuous}
James Harrison, Apoorva Sharma, Chelsea Finn, and Marco Pavone.
\newblock Continuous meta-learning without tasks.
\newblock {\em Neural Information Processing Systems (NeurIPS)}, 2020.

\bibitem{harrison2018meta}
James Harrison, Apoorva Sharma, and Marco Pavone.
\newblock Meta-learning priors for efficient online {Bayesian} regression.
\newblock {\em Workshop on the Algorithmic Foundations of Robotics (WAFR)},
  2018.

\bibitem{he2016deep}
Kaiming He, Xiangyu Zhang, Shaoqing Ren, and Jian Sun.
\newblock Deep residual learning for image recognition.
\newblock {\em {IEEE} Conference on Computer Vision and Pattern Recognition
  (CVPR)}, 2016.

\bibitem{hochreiter2001learning}
Sepp Hochreiter, A~Steven Younger, and Peter~R Conwell.
\newblock Learning to learn using gradient descent.
\newblock {\em International Conference on Artificial Neural Networks}, 2001.

\bibitem{hospedales2020meta}
Timothy Hospedales, Antreas Antoniou, Paul Micaelli, and Amos Storkey.
\newblock Meta-learning in neural networks: A survey.
\newblock {\em arXiv:2004.05439}, 2020.

\bibitem{jerfel2019reconciling}
Ghassen Jerfel, Erin Grant, Tom Griffiths, and Katherine~A Heller.
\newblock Reconciling meta-learning and continual learning with online mixtures
  of tasks.
\newblock {\em Neural Information Processing Systems (NeurIPS)}, 2019.

\bibitem{jordan2015gentle}
Michael Jordan and Yee~Whye Teh.
\newblock A gentle introduction to the dirichlet process, the beta process and
  bayesian nonparametrics, 2015.

\bibitem{jordan1995logistic}
Michael~I Jordan.
\newblock Why the logistic function? a tutorial discussion on probabilities and
  neural networks.
\newblock {\em Computational Cognitive Science Report}, 1995.

\bibitem{kirkpatrick2017overcoming}
James Kirkpatrick, Razvan Pascanu, Neil Rabinowitz, Joel Veness, Guillaume
  Desjardins, Andrei~A Rusu, Kieran Milan, John Quan, Tiago Ramalho, Agnieszka
  Grabska-Barwinska, et~al.
\newblock Overcoming catastrophic forgetting in neural networks.
\newblock {\em Proceedings of the national academy of sciences}, 2017.

\bibitem{lee2018simple}
Kimin Lee, Kibok Lee, Honglak Lee, and Jinwoo Shin.
\newblock A simple unified framework for detecting out-of-distribution samples
  and adversarial attacks.
\newblock {\em Neural Information Processing Systems (NeurIPS)}, 2018.

\bibitem{lee2019meta}
Kwonjoon Lee, Subhransu Maji, Avinash Ravichandran, and Stefano Soatto.
\newblock Meta-learning with differentiable convex optimization.
\newblock {\em {IEEE} Conference on Computer Vision and Pattern Recognition
  (CVPR)}, 2019.

\bibitem{liu2020openset}
Bo Liu, Hao Kang, Haoxiang Li, Gang Hua, and Nuno Vasconcelos.
\newblock Few-shot open-set recognition using meta-learning.
\newblock {\em {IEEE} Conference on Computer Vision and Pattern Recognition
  (CVPR)}, 2020.

\bibitem{murphy2012machine}
Kevin~P Murphy.
\newblock {\em Machine Learning: A Probabilistic Perspective}.
\newblock MIT Press, 2012.

\bibitem{nagabandi2019deep}
Anusha Nagabandi, Chelsea Finn, and Sergey Levine.
\newblock Deep online learning via meta-learning: Continual adaptation for
  model-based {RL}.
\newblock {\em arXiv:1812.07671}, 2019.

\bibitem{neal2018open}
Lawrence Neal, Matthew Olson, Xiaoli Fern, Weng-Keen Wong, and Fuxin Li.
\newblock Open set learning with counterfactual images.
\newblock {\em European Conference on Computer Vision (ECCV)}, 2018.

\bibitem{neal2000markov}
Radford~M Neal.
\newblock Markov chain sampling methods for dirichlet process mixture models.
\newblock {\em Journal of computational and graphical statistics}, 2000.

\bibitem{orbanz2010bayesian}
Peter Orbanz and Yee~Whye Teh.
\newblock Bayesian nonparametric models.
\newblock {\em Encyclopedia of machine learning}, 2010.

\bibitem{rajeswaran2019meta}
Aravind Rajeswaran, Chelsea Finn, Sham~M Kakade, and Sergey Levine.
\newblock Meta-learning with implicit gradients.
\newblock {\em Neural Information Processing Systems (NeurIPS)}, 2019.

\bibitem{ravi2017optimization}
Sachin Ravi and Hugo Larochelle.
\newblock Optimization as a model for few-shot learning.
\newblock {\em International Conference on Learning Representations (ICLR)},
  2017.

\bibitem{ravichandran2020incremental}
Avinash Ravichandran, Rahul Bhotika, and Stefano Soatto.
\newblock Incremental few-shot meta-learning via indirect discriminant
  alignment.
\newblock {\em European Conference on Computer Vision (ECCV)}, 2020.

\bibitem{rebuffi2017icarl}
Sylvestre-Alvise Rebuffi, Alexander Kolesnikov, Georg Sperl, and Christoph~H
  Lampert.
\newblock icarl: Incremental classifier and representation learning.
\newblock {\em {IEEE} Conference on Computer Vision and Pattern Recognition
  (CVPR)}, 2017.

\bibitem{ren2020wandering}
Mengye Ren, Michael~L Iuzzolino, Michael~C Mozer, and Richard~S Zemel.
\newblock Wandering within a world: Online contextualized few-shot learning.
\newblock {\em arXiv preprint arXiv:2007.04546}, 2020.

\bibitem{ren2019attractor}
Mengye Ren, Renjie Liao, Ethan Fetaya, and Richard Zemel.
\newblock Incremental few-shot learning with attention attractor networks.
\newblock {\em Neural Information Processing Systems (NeurIPS)}, 2019.

\bibitem{ren2018meta}
Mengye Ren, Eleni Triantafillou, Sachin Ravi, Jake Snell, Kevin Swersky,
  Joshua~B Tenenbaum, Hugo Larochelle, and Richard~S Zemel.
\newblock Meta-learning for semi-supervised few-shot classification.
\newblock {\em International Conference on Learning Representations (ICLR)},
  2018.

\bibitem{robbins1956empirical}
Herbert Robbins.
\newblock An empirical {B}ayes approach to statistics.
\newblock {\em Third Berkeley symposium on Mathematical statistics and
  Probability}, 1956.

\bibitem{rudd2018evm}
Ethan~M. Rudd, Lalit~P. Jain, Walter~J. Scheirer, and Terrance~E. Boult.
\newblock The extreme value machine.
\newblock {\em IEEE Transactions on Pattern Analysis \& Machine Intelligence},
  2018.

\bibitem{santoro2016meta}
Adam Santoro, Sergey Bartunov, Matthew Botvinick, Daan Wierstra, and Timothy
  Lillicrap.
\newblock Meta-learning with memory-augmented neural networks.
\newblock {\em International Conference on Machine Learning (ICML)}, 2016.

\bibitem{snell2017prototypical}
Jake Snell, Kevin Swersky, and Richard Zemel.
\newblock Prototypical networks for few-shot learning.
\newblock {\em Neural Information Processing Systems (NeurIPS)}, 2017.

\bibitem{triantafillou2019meta}
Eleni Triantafillou, Tyler Zhu, Vincent Dumoulin, Pascal Lamblin, Utku Evci,
  Kelvin Xu, Ross Goroshin, Carles Gelada, Kevin Swersky, Pierre-Antoine
  Manzagol, et~al.
\newblock Meta-dataset: A dataset of datasets for learning to learn from few
  examples.
\newblock {\em International Conference on Learning Representations (ICLR)},
  2019.

\bibitem{vanschoren2018meta}
Joaquin Vanschoren.
\newblock Meta-learning: A survey.
\newblock {\em arXiv:1810.03548}, 2018.

\bibitem{vinyals2016matching}
Oriol Vinyals, Charles Blundell, Timothy Lillicrap, Koray Kavukcuoglu, and Daan
  Wierstra.
\newblock Matching networks for one shot learning.
\newblock {\em Neural Information Processing Systems (NeurIPS)}, 2016.

\bibitem{wallingford2020wild}
Matthew Wallingford, Aditya Kusupati, Keivan Alizadeh-Vahid, Aaron Walsman,
  Aniruddha Kembhavi, and Ali Farhadi.
\newblock In the wild: From ml models to pragmatic ml systems.
\newblock {\em arXiv:2007.02519}, 2020.

\bibitem{ye2021learning}
Han-Jia Ye, Hexiang Hu, and De-Chuan Zhan.
\newblock Learning adaptive classifiers synthesis for generalized few-shot
  learning.
\newblock {\em International Journal of Computer Vision}, 129(6):1930--1953,
  2021.

\end{thebibliography}
\bibliographystyle{ieee_fullname}

}

%

\begin{IEEEbiography}[{\includegraphics[width=1in,height=1.25in,clip,keepaspectratio]{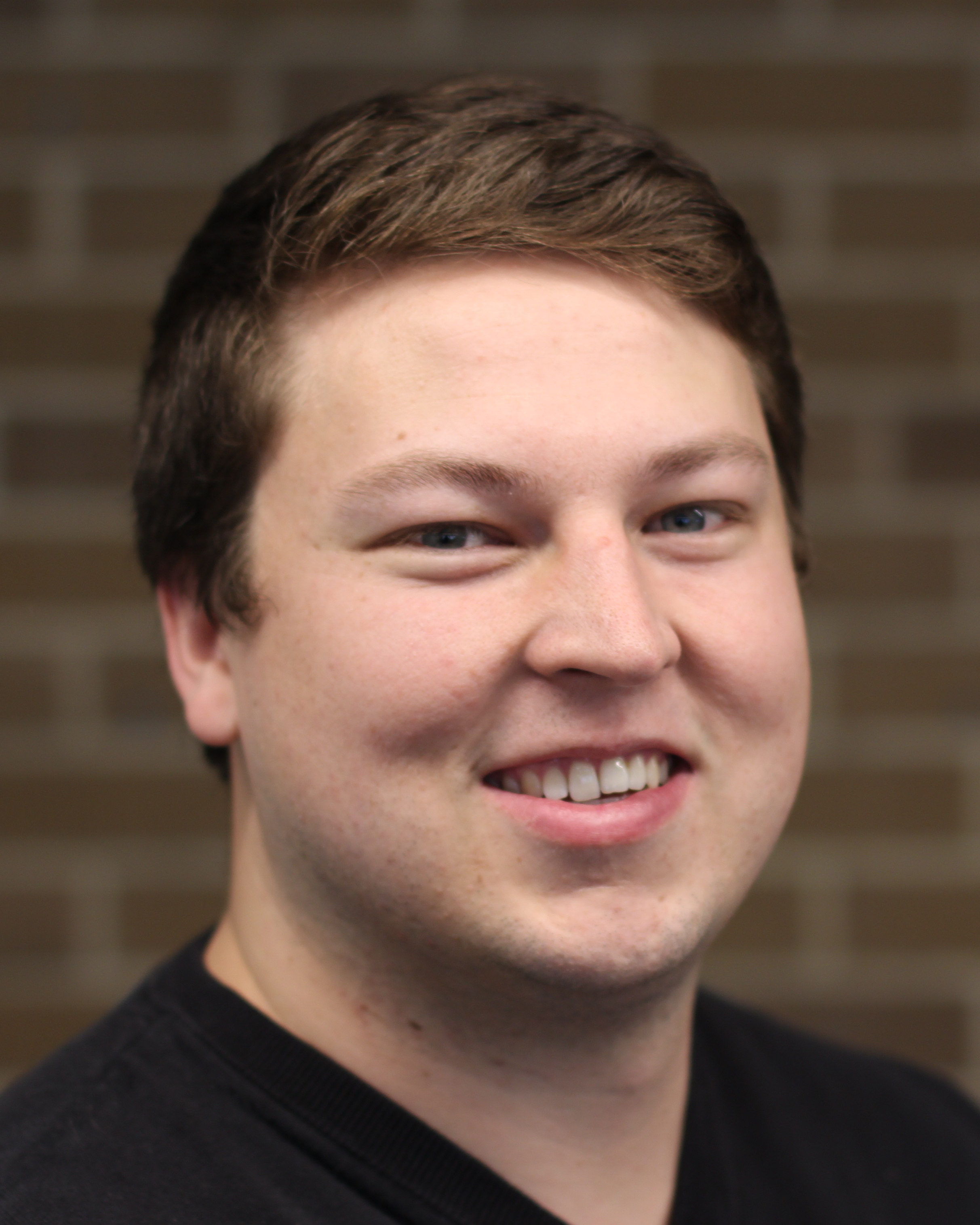}}]{John Willes}
is currently a MASc candidate at The University of Toronto Institute for Aerospace Studies (UTIAS), Toronto, Canada. He received a B.Eng.~degree in Mechanical Engineering from McGill University in 2015. His research is currently focused on few-shot learning and 3D object detection and tracking for autonomous driving and intelligent transportation systems.
\end{IEEEbiography}

\begin{IEEEbiography}[{\includegraphics[width=1in,height=1.25in,clip,keepaspectratio]{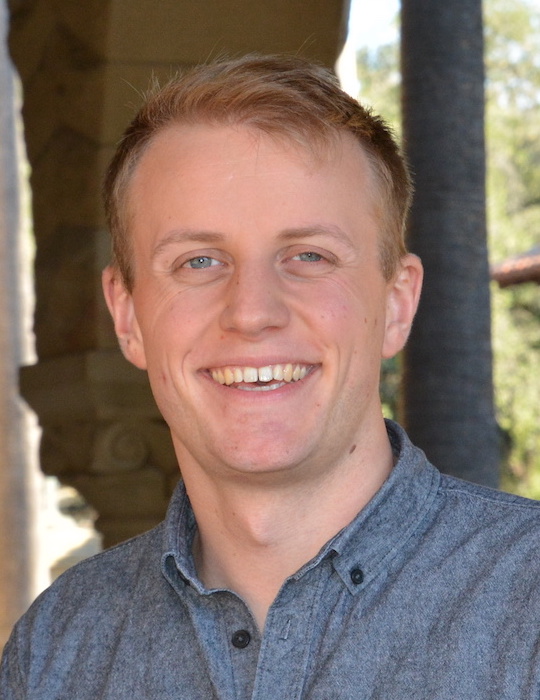}}]{James Harrison}
is a Ph.D.~candidate in the Autonomous Systems Lab at Stanford University. He received an M.S.~degree from Stanford University in 2018 and a B.Eng.~degree from McGill University in 2015, both in mechanical engineering. His research interests include few-shot, adaptive, and open-world learning, Bayesian deep learning, and applications in safe robot autonomy, decision-making, and control. 
\end{IEEEbiography}

\begin{IEEEbiography}[{\includegraphics[width=1in,height=1.25in,clip,keepaspectratio]{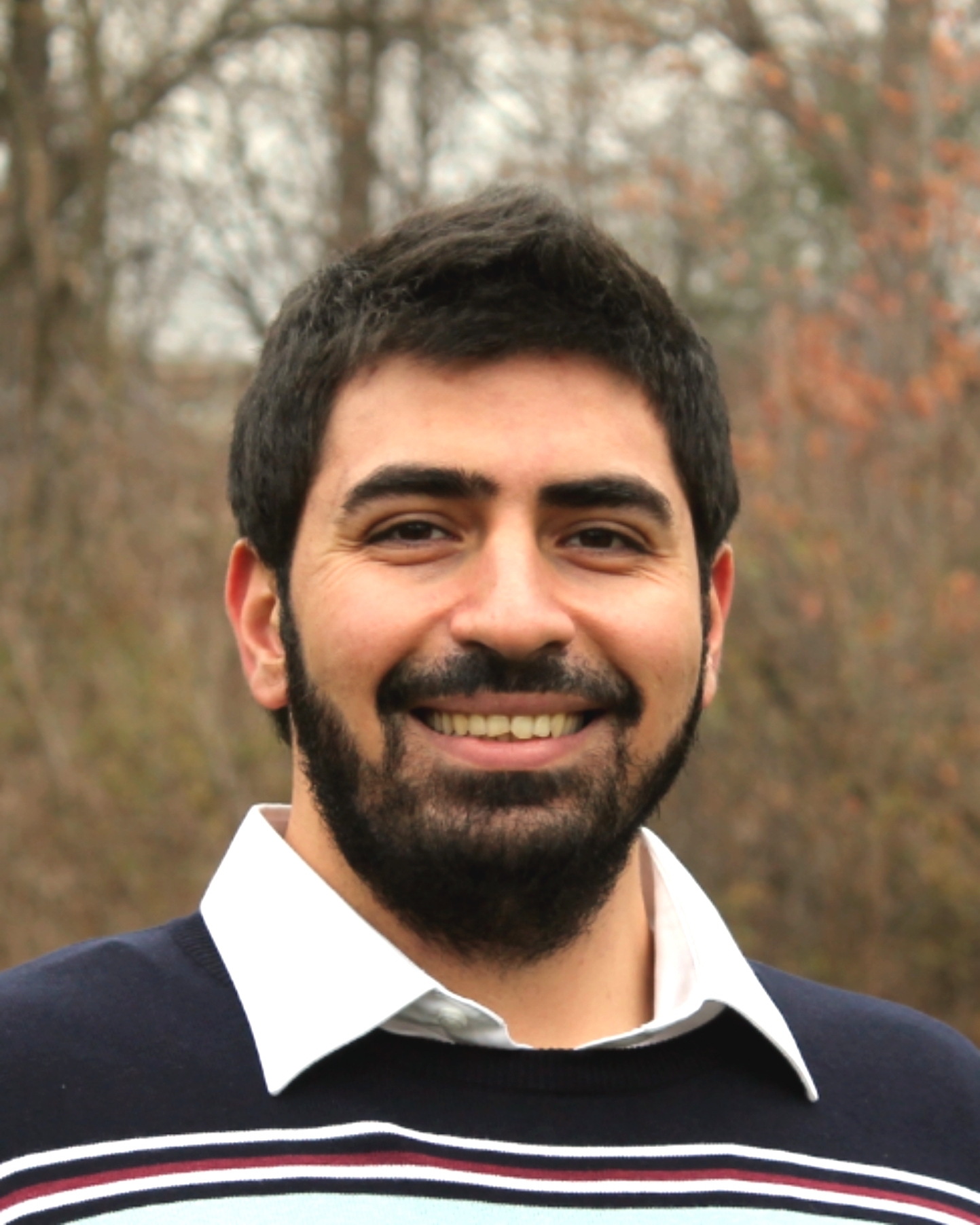}}]{Ali Harakeh} (Member, IEEE) is currently PhD candidate at The University of Toronto Institute for Aerospace Studies (UTIAS), Toronto, Canada.  During his studies, he developed real-time perception algorithms that were part of autonomous driving live demonstrations in the Consumer Electronics Show (CES) and The Vehicular Technology Conference (VTC) in 2017. He also worked on developing cutting edge perception algorithms in cooperation with companies such as Huawei Technologies, Renesas Electronics, and LG Electronics. He received his master's degree in mechanical engineering from The American University of Beirut, Beirut, Lebanon in 2016. He currently specializes in perception algorithms for autonomous driving, working on fundamental research within 2D and 3D object detection, semantic segmentation, uncertainty estimation for deep neural networks, and synthetic data generation.
\end{IEEEbiography}

\begin{IEEEbiography}[{\includegraphics[width=1in,height=1.25in,clip,keepaspectratio]{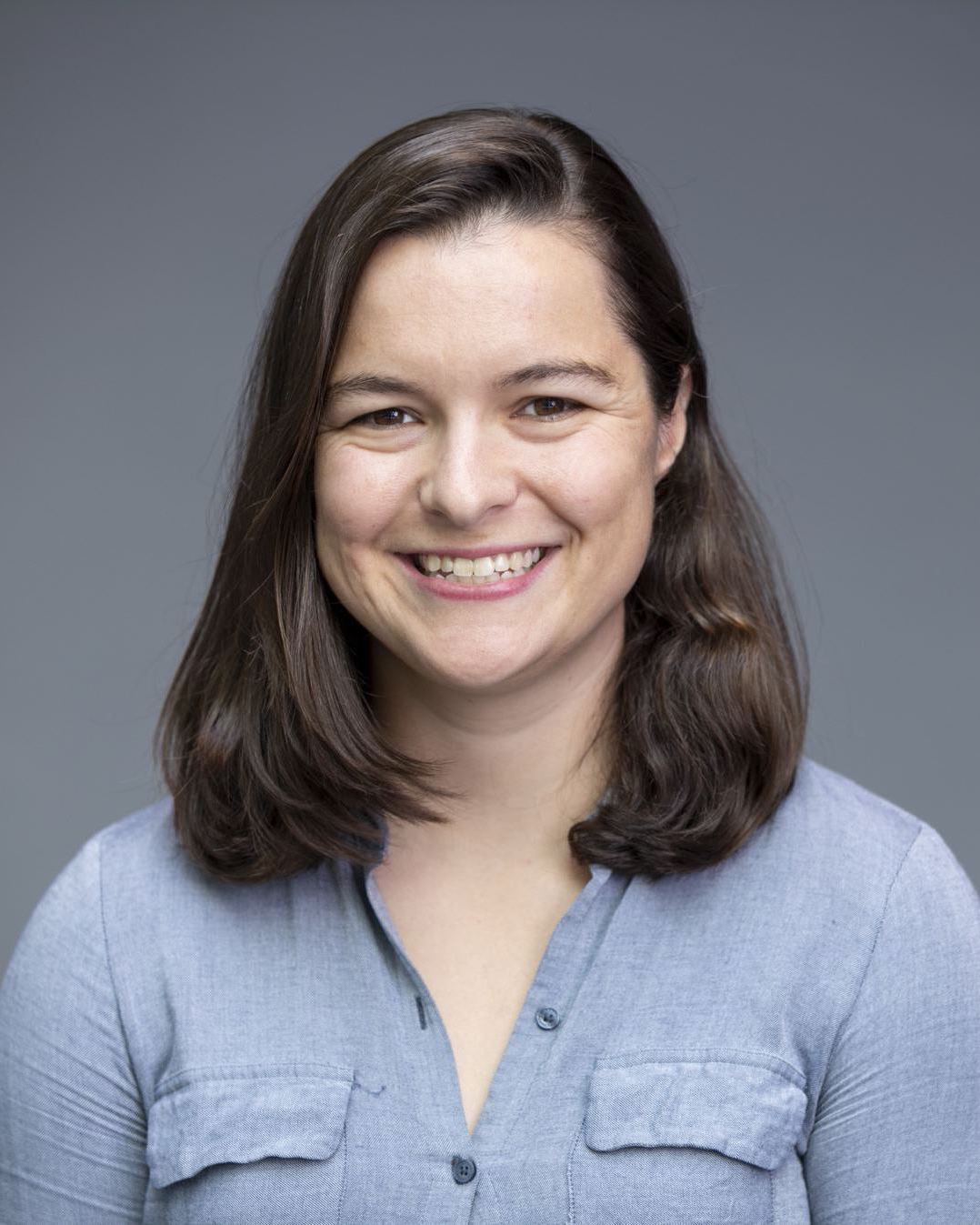}}]{Chelsea Finn} is an Assistant Professor in Computer Science and Electrical Engineering at Stanford University. Her research interests lie in the capability of robots and other agents to develop broadly intelligent behavior through learning and interaction. She received her Bachelor's degree in Electrical Engineering and Computer Science at MIT and her PhD in Computer Science at UC Berkeley. Her research has been recognized through the ACM doctoral dissertation award, the Microsoft Research Faculty Fellowship, the C.V. Ramamoorthy Distinguished Research Award, and the MIT Technology Review 35 under 35 Award, and her work has been covered by various media outlets, including the New York Times, Wired, and Bloomberg.
\end{IEEEbiography}

\begin{IEEEbiography}[{\includegraphics[width=1in,height=1.25in,clip,keepaspectratio]{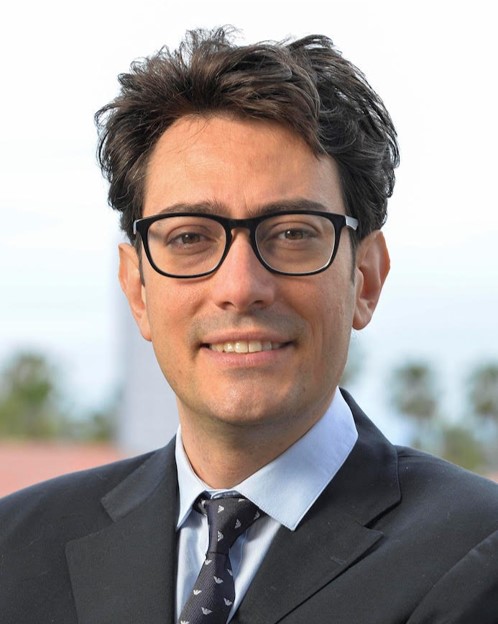}}]{Marco Pavone}
is an Associate Professor of Aeronautics and Astronautics at Stanford University, where he is the Director of the Autonomous Systems Laboratory.
He received a Ph.D. degree in Aeronautics and Astronautics from MIT in 2010. His main research interests are in the development of methodologies for the analysis, design, and control of autonomous systems, with an emphasis on self-driving cars, autonomous aerospace vehicles, and future mobility systems. 
He is currently  an Associate Editor for the IEEE Control Systems Magazine.
\end{IEEEbiography}

\begin{IEEEbiography}[{\includegraphics[width=1in,height=1.25in,clip,keepaspectratio]{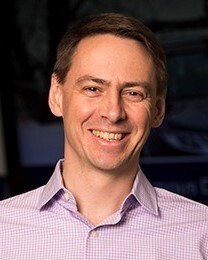}}]{Steven L. Waslander} (Senior Member, IEEE) is a leading authority on autonomous aerial and ground vehicles, including multirotor drones and autonomous driving vehicles, Simultaneous Localization and Mapping (SLAM) and multi-vehicle systems.  He received his B.Sc.E.~in 1998 from Queen’s University, his M.S.~in 2002 and his Ph.D.~in 2007, both from Stanford University in Aeronautics and Astronautics. He joined the University of Waterloo in 2008, where he founded and directed the Waterloo Autonomous Vehicle Laboratory (WAVELab). In 2018, he joined the University of Toronto Institute for Aerospace Studies (UTIAS), and founded the Toronto Robotics and Artificial Intelligence Laboratory (TRAILab). Prof. Waslander’s innovations were recognized by the Ontario Centres of Excellence Mind to Market award for the best Industry/Academia collaboration (2012, with Aeryon Labs), as well as best paper and best poster awards at the Computer and Robot Vision Conference (2018).  His work on autonomous vehicles has resulted in the Autonomoose, the first autonomous vehicle created at a Canadian University to drive on public roads.
\end{IEEEbiography}



\vfill


\end{document}